\documentclass[runningheads]{llncs}

\usepackage{eccv}
\usepackage{eccvabbrv}
\usepackage{microtype}
\usepackage{amsmath,amsfonts}
\usepackage{textcomp}
\usepackage{siunitx}
\usepackage{amsmath,amsfonts,bm}

\def\eqref#1{equation~\ref{#1}}
\def\1{\bm{1}}

\DeclareMathAlphabet{\mathsfit}{\encodingdefault}{\sfdefault}{m}{sl}
\SetMathAlphabet{\mathsfit}{bold}{\encodingdefault}{\sfdefault}{bx}{n}

\makeatletter
\DeclareRobustCommand\onedot{\futurelet\@let@token\@onedot}
\def\@onedot{\ifx\@let@token.\else.\null\fi\xspace}
 
\def\ie{i.e\onedot}

\def\etal{et~al\onedot}

\makeatother

\usepackage{overpic}
\usepackage{enumitem} %< control spacing in itemize/enumerate/...
\usepackage{overpic} %< add raw math symbols to figures
\usepackage{color}
\usepackage{mathtools} %< \coloneqq
\usepackage{currfile}
\usepackage{multirow}
\definecolor{turquoise}{cmyk}{0.65,0,0.1,0.3}
\definecolor{purple}{rgb}{0.65,0,0.65}
\definecolor{dark_green}{rgb}{0, 0.5, 0}
\definecolor{orange}{rgb}{0.8, 0.6, 0.2}
\definecolor{red}{rgb}{0.8, 0.2, 0.2}
\definecolor{darkred}{rgb}{0.6, 0.1, 0.05}
\definecolor{blueish}{rgb}{0.0, 0.3, .6}
\definecolor{light_gray}{rgb}{0.7, 0.7, .7}
\definecolor{pink}{rgb}{1, 0, 1}
\definecolor{greyblue}{rgb}{0.25, 0.25, 1}

\renewcommand{\paragraph}[1]{\vspace{.5em}\noindent\textbf{#1}.}
\usepackage{lipsum}

\usepackage{blindtext}

\newcommand{\kostas}[1]{{\color{Bittersweet} {[\bf Kosta: #1]}}}
\newcommand{\andrew}[1]{{\color{BlueViolet} {[Andrew: #1]}}}
\newcommand{\vitto}[1]{{\color{OrangeRed} {[Vitto: #1]}}}
\newcommand{\JB}[1]{{\color{OliveGreen} {[Jon: #1]}}}
\newcommand{\tom}[1]{{\color{RoyalPurple} {[Tom: #1]}}}
\newcommand{\pratul}[1]{{\color{Emerald} {[Pratul: #1]}}}
\newcommand{\at}[1]{{\color{blueish}#1}}
\newcommand{\AT}[1]{{\color{blueish}{\bf [Andrea: #1]}}}
\newcommand{\At}[1]{\marginpar{\tiny{\textcolor{blueish}{#1}}}}
\newcommand{\al}[1]{\textbf{\color{orange}[AL: #1]}}
\renewcommand{\kostas}[1]{}
\renewcommand{\andrew}[1]{}
\renewcommand{\vitto}[1]{}
\renewcommand{\JB}[1]{}
\renewcommand{\tom}[1]{}
\renewcommand{\pratul}[1]{}
\renewcommand{\at}[1]{}
\renewcommand{\AT}[1]{}
\renewcommand{\At}[1]{}
\renewcommand{\al}[1]{}

\usepackage{algorithm}
\usepackage{algpseudocode}

\usepackage{graphicx}
\usepackage{wrapfig}
\usepackage{float}
\usepackage{stfloats}
\usepackage{booktabs}
\usepackage{multirow}
\usepackage{makecell}
\usepackage{array}
\usepackage{hhline}
\usepackage{color}
\usepackage[table]{xcolor}

\usepackage{verbatim}
\usepackage{comment}
\usepackage{cite}
\usepackage{url}
\usepackage[accsupp]{axessibility}

\usepackage[pagebackref,breaklinks,colorlinks,citecolor=eccvblue]{hyperref}
\usepackage{orcidlink}

\begin{document}

\title{PRISM3D: Probabilistic Refinement and Robust Initialization for Physically Consistent Scene Modeling under Extreme Motion Blur \vspace{-4mm}} 
\titlerunning{PRISM3D: Robust 3D Scene Modeling under Extreme Motion Blur}

\author{Gopi Raju Matta\orcidlink{0000-0003-3305-3136} \and
Reddypalli Trisha\orcidlink{0000-0005-2181-8324} \and
Vemunuri Divya Madhuri\orcidlink{0009-0007-2876-8512} \and
Kaushik Mitra\orcidlink{0000-0001-6747-9050}}

\authorrunning{G.~R.~Matta et al.}

\institute{Computational Imaging Lab, Dept. of Electrical Engineering, IIT Madras\\
\email{\{ee17d021, ee26s067, ee24d004\}@smail.iitm.ac.in, kmitra@ee.iitm.ac.in}}

\maketitle

\begin{abstract}
\vspace{-8mm}
We address the inverse problem of \textbf{blind 3D scene reconstruction} from extremely motion-blurred images, a scenario where traditional Structure-from-Motion (SfM) pipelines fail. Existing approaches typically circumvent this bottleneck by relying on impractical sharp-image supervision. In this work, we introduce \textbf{PRISM3D}, a unified framework enabling robust reconstruction directly from severely degraded inputs. To overcome the lack of a reliable starting point, we propose a \textbf{Robust Initialization} strategy utilizing deep dense tracking method (VGGSfM) to recover global topology where feature matching fails. To the best of our knowledge, we are the first to effectively leverage this paradigm to bootstrap 3D Gaussian Splatting from extreme motion blur. However, while robust, this initialization yields sparse and noisy geometry that causes deterministic optimization to diverge. To resolve this, we propose a coupled solution driven by probability and physics: we \textbf{adopt a probabilistic formulation} for geometric densification via Markov Chain Monte Carlo (MCMC) to robustly populate the sparse priors, while simultaneously modeling physical image formation via continuous \textbf{Bézier Trajectories}. Furthermore, while PRISM3D establishes a highly robust standalone pipeline, the availability of complementary event streams offers an opportunity to push the reconstruction fidelity further. To exploit this, we introduce \textbf{PRISM3D-E}, a multi-modal (RGB + Events) extension that seamlessly integrates high-temporal-resolution events as structural priors to maximize geometric recovery. Because existing datasets lack paired event streams under such severe degradation, we concurrently contribute the \textbf{PRISM3D-E Benchmark} to facilitate rigorous evaluation. Extensive experiments demonstrate that both our standalone RGB framework and its multi-modal extension establish new state-of-the-art performance. To facilitate future research in extreme motion deblurring, our project page, featuring extensive qualitative comparisons, video results, and the upcoming releases of our codebase and benchmark dataset, is available at \url{https://gopirajumatta.github.io/PRISM3D/}.
\vspace{-6mm}
\end{abstract}

\begin{comment}
\keywords{3D Gaussian Splatting \and Motion Deblurring \and Novel View Synthesis \and Deep Structure-from-Motion \and Probabilistic Modeling \and Event Cameras}
\end{comment}

\section{Introduction}
\vspace{-4mm}
Recovering high-fidelity 3D scenes from extremely motion-blurred images is a persistent inverse problem in computer vision. In real-world scenarios like robotics or handheld photography, rapid camera motion inevitably degrades image quality. Computational reconstruction requires solving a fundamental ``chicken-and-egg'' paradox: accurate deblurring requires precise camera poses, yet estimating poses from blurred images requires a deblurred scene model to establish correspondences.

Despite the rapid evolution of Neural Radiance Fields (NeRF)~\cite{mildenhall2020nerf} and 3D Gaussian Splatting (3DGS)~\cite{kerbl3Dgaussians}, existing frameworks largely ignore the first half of this paradox. State-of-the-art methods typically rely on an ``oracle'' assumption: that accurate camera poses and point clouds are provided by Structure-from-Motion (SfM) pipelines like COLMAP~\cite{colmap} running on sharp images. This renders them impractical for blind deblurring. As demonstrated in our systematic analysis (Sec.~\ref{sec:robustness_blur_levels}), traditional SfM pipelines are brittle under extreme blur. Because blur acts as a low-pass filter, it obliterates the high-frequency features (e.g., corners, edges) required to establish reliable correspondences. Consequently, at moderate-to-extreme blur levels, standard initialization fails completely.

{Recent advancements offer alternative paths to bypass this registration failure. Deep Structure-from-Motion (SfM) pipelines like VGGSfM~\cite{wang2024vggsfm} can successfully recover global topology directly from degraded inputs. Concurrently, feed-forward 3D Foundation Models (3DFMs) like VGGT~\cite{vggt} provide rapid scene initialization by directly predicting dense geometry. However, these dense but highly noisy point clouds trap initialization-sensitive 3DGS training in local minima~\cite{vggt-x}. While Deep SfM avoids these local minima by extracting a geometrically consistent, albeit sparse, topological skeleton, using it as a simple drop-in replacement for COLMAP remains insufficient. Its inherent geometric sparsity leads to inferior, artifact-prone reconstructions when paired with standard deterministic 3DGS optimization.}

{To bridge this \textbf{Initialization-Optimization Gap}, we introduce \textbf{PRISM3D}. We formulate blind deblurring as a joint optimization problem that explicitly addresses both the uncertainty of this deep blind initialization and the continuous dynamics of camera motion. Specifically, we tightly couple robust deep initialization with probabilistic scene densification (via MCMC) to gracefully populate the sparse geometry, while effectively capturing the physical blur image formation process using continuous Bézier trajectories. This yields a highly robust pipeline comprising three theoretically grounded components:}

\textbf{Robust Initialization via Deep Topology Recovery:}
We eliminate the dependency on brittle feature matching by leveraging VGGSfM~\cite{wang2024vggsfm}. Through a controlled evaluation of blur severity (1 to 11 frames), we demonstrate that VGGSfM maintains stable registration where COLMAP fails, providing the essential topological skeleton required to bootstrap optimization.

\textbf{Probabilistic Scene Modeling (The MCMC Formulation):}
To handle the high variance of our deep initialization, we abandon heuristic densification in favor of a probabilistic formulation using 3DGS-MCMC~\cite{MCMC}. We treat Gaussian primitives as samples from an underlying scene distribution rather than fixed parameters. By updating these samples via Stochastic Gradient Langevin Dynamics (SGLD), our framework robustly explores the solution space and ``fills in'' structural gaps, critically preventing local minima.

\textbf{Physics-Consistent Trajectory Optimization:}
We solve the inverse image formation problem by modeling camera motion as a continuous $SE(3)$ Bézier trajectory. By integrating the scene radiance along this curved path, we mathematically formulate the blur as a differentiable function of the camera's momentum. Jointly optimizing these trajectories with the probabilistic scene model ensures that the reconstruction is physically consistent with the exposure process~\cite{zhao2024badgaussians}.

\textbf{Multi-Modal Degenerate Recovery:}
While PRISM3D is a highly robust standalone pipeline, the limits of reconstruction fidelity can be pushed further when complementary event sensor data is available. We introduce \textbf{PRISM3D-E}, a multi-modal (RGB + Events) extension designed for completely degenerate scenarios where photometric cues are nearly obliterated. PRISM3D-E leverages the microsecond resolution of event cameras via the Event-based Double Integral (EDI) model~\cite{pan2019bringing} to extract high-frequency structural priors during the exposure window. Routing these priors through our Deep SfM pipeline guarantees a strictly superior topological skeleton to bootstrap our MCMC densification. To facilitate rigorous evaluation of this extreme regime, we concurrently contribute the \textbf{PRISM3D-E Benchmark}.

Our contributions are as follows:
\begin{itemize}
    \item We propose \textbf{PRISM3D}, a unified framework for solving blind deblurring by coupling \textbf{Robust Deep Initialization} (VGGSfM) and \textbf{Probabilistic Refinement} (MCMC) with continuous \textbf{Bézier Trajectories}. To our knowledge, this is the first work to effectively bootstrap 3DGS from extreme motion blur without sharp image supervision.
   \item {We present a \textbf{systematic analysis of robustness} across varying blur levels, empirically validating the necessity of each core component. Specifically, we demonstrate that \textbf{Deep SfM} maintains reliable initialization where classical COLMAP fails, that our \textbf{Probabilistic MCMC} formulation is critical for resolving the geometric sparsity of deep tracking, and that continuous \textbf{Bézier trajectories} strictly outperform linear and cubic splines in modeling complex camera motion.}
   \item {We introduce the \textbf{PRISM3D-E Benchmark}, comprising a novel dataset that explicitly pairs extreme motion blur with complementary event streams, supported by standardized evaluation protocols and multi-modal baselines. Alongside this, we propose the \textbf{PRISM3D-E} multi-modal extension, demonstrating that routing event-derived priors through deep tracking can \textbf{further enhance reconstruction} in completely degenerate scenarios.}
    \item Through extensive evaluations, we demonstrate that both our standalone framework (\textbf{PRISM3D}) and its multi-modal extension (\textbf{PRISM3D-E}) achieve state-of-the-art reconstruction quality, effectively bridging the gap between deep structure estimation and \textbf{physics-based differentiable rendering}.
\end{itemize}

\vspace{-6mm}

\section{Related Work}
\label{sec:related_work}
\vspace{-2mm}
\subsection{Robust Initialization in Degenerate Scenarios}
Accurate pose estimation is a prerequisite for optimizing NeRF and 3D Gaussian Splatting (3DGS). However, standard SfM pipelines like COLMAP~\cite{colmap} and its enhanced variants (GLOMAP~\cite{glowmap}, HLOC~\cite{hloc} with SuperGlue~\cite{superglue}, Pixel-Perfect SfM~\cite{pixel_perfect_sfm}) rely on local feature correspondences, making them notoriously brittle under extreme motion blur. 

Feed-forward 3D Foundation Models (3DFMs) like VGGT~\cite{vggt} and VGGT-X~\cite{vggt-x} offer rapid initialization, but scaling them to dense views yields noisy dense point clouds that trap initialization-sensitive 3DGS training in local minima. Conversely, deep dense tracking provides a geometrically rigorous alternative. By replacing discrete keypoint matching with differentiable tracking and bundle adjustment~\cite{co3d,phototourism,eth3d}, VGGSfM~\cite{wang2024vggsfm} serves as a critical enabler for \textit{blind} deblurring, recovering global topology where classical SfM and 3DFMs fail.
\vspace{-4mm}

\subsection{Probabilistic Scene Representation}
While NeRF~\cite{mildenhall2020nerf} and its implicit variants~\cite{barron2021mip,barron2022mip,jiang2022alignerf,wu2021diver,kaizhang2020nerfplusplus,lindell2021autoint,Reiser2021ICCV, yu2021plenoctrees,fridovich2022plenoxels,muller2022instant,chen2022tensorf,sun2022dvgo} revolutionized view synthesis, they suffer from computationally expensive ray marching~\cite{maxraymarching}. \textit{3D Gaussian Splatting} (3DGS)~\cite{kerbl20233d} resolves this via efficient rasterization. However, standard 3DGS uses deterministic cloning and pruning heuristics assuming a dense, low-variance initialization (e.g., from COLMAP). Under the sparse initializations common in blind deblurring, these heuristics fail to meaningfully populate the scene, causing geometric collapse.

\textit{3DGS-MCMC}~\cite{MCMC} addresses this by reformulating Gaussians as probabilistic samples from a scene distribution, utilizing Markov Chain Monte Carlo for robust densification. We adopt this formulation to dynamically populate the sparse geometry from our deep SfM initialization, bridging the gap between noisy tracking and high-fidelity rendering.
\vspace{-4mm}

\subsection{Blind 3D Deblurring}
Early NeRF-based deblurring methods like BAD-NeRF~\cite{wang2023badnerf} converge slowly. While explicit 3DGS representations overcome this bottleneck, they remain limited by initialization strategies. ExBluRF~\cite{lee2023exblurf} and Deblur-GS~\cite{chen2024deblur} assume impractical access to sharp reference images. Conversely, blind methods like BAD-Gaussians~\cite{zhao2024badgaussians} attempt physical trajectory modeling but critically rely on standard COLMAP, failing on extremely blurred inputs.

To bypass photometric degradation, multi-modal approaches use supplementary hardware. Jang~\etal~\cite{jang2025splat} require active RGB-D sensors. Event-based NeRFs~\cite{qi2023e2nerf,ebadnerf} inherit implicit network bottlenecks, while recent event-assisted 3DGS methods (EvaGaussians~\cite{yu2025evagaussians}, E2GS~\cite{e2gs}) synthesize pseudo-sharp images purely to facilitate COLMAP initialization. Thus, they remain inapplicable in standard RGB-only scenarios and bottlenecked by classical SfM. In contrast, \textbf{PRISM3D} is an \textit{event-agnostic}, standalone RGB solution capable of surviving extreme degradation. When complementary events \textit{are} available, our multi-modal extension (\textbf{PRISM3D-E}) seamlessly integrates them to yield strictly superior geometry over SfM-dependent baselines.

\begin{figure}[t]
    \centering
    \includegraphics[width=0.98\linewidth]{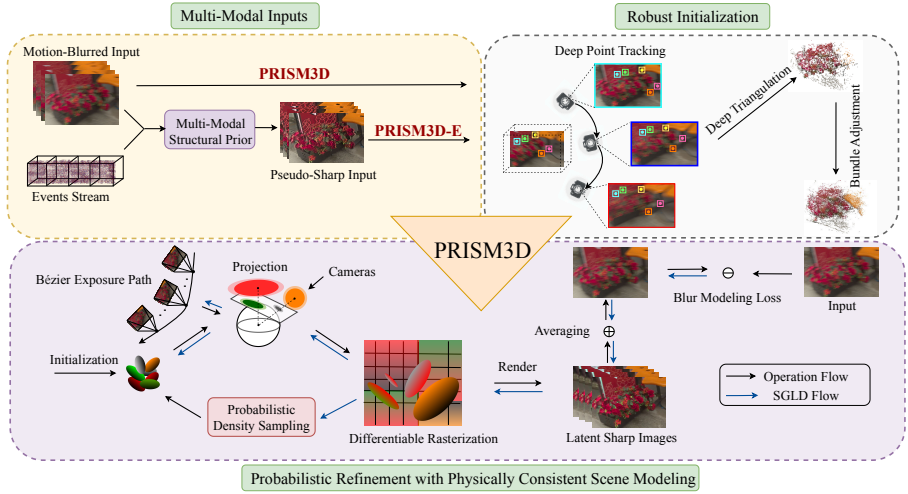}
    \caption{\textbf{Overview of the PRISM3D Framework.} We address extreme motion blur through a unified inverse problem formulation. \textbf{(Top Left)} For the standalone \textbf{PRISM3D} pipeline, we input blurry images directly. For \textbf{PRISM3D-E}, we incorporate complementary event streams to generate high-frequency structural priors (EDI). \textbf{(Top Right)} These inputs feed into our \textbf{Robust Initialization} stage, which utilizes deep point tracking and bundle adjustment to recover a reliable sparse point cloud and camera poses, bypassing brittle classical feature matching. \textbf{(Bottom)} This sparse geometry bootstraps our \textbf{Probabilistic Refinement} module. We model the scene using 3D Gaussians updated via Stochastic Gradient Langevin Dynamics (SGLD flow, blue arrows) to robustly populate structural gaps. Concurrently, we ensure physical consistency by rendering latent sharp views along a continuous \textbf{Bézier Exposure Path}, averaging them to compute the blur modeling loss against the degraded input.}
    \label{fig:method-pipeline}
    \vspace{-6mm}
\end{figure}
\vspace{-4mm}
\section{Method}
\label{sec:method}
\vspace{-2mm}
\textbf{PRISM3D} is a unified 3D Gaussian Splatting (3DGS) framework designed for robust novel view synthesis under extreme motion blur. To bridge the initialization-optimization gap, our architecture replaces brittle feature matching with deep dense tracking via VGGSfM~\cite{wang2024vggsfm}. Because this deep initialization is inherently sparse, we couple it with a probabilistic MCMC-based densification strategy~\cite{MCMC}. Finally, to ensure physical accuracy, we jointly optimize the scene geometry alongside continuous camera trajectories. Furthermore, we introduce \textbf{PRISM3D-E}, a multi-modal extension that leverages complementary event data, when available, to extract high-frequency structural priors and push reconstruction limits even further. Our complete framework is illustrated in Figure~\ref{fig:method-pipeline}.

\vspace{-4mm}
\subsection{PRISM3D: Blind RGB Extreme Motion Deblurring}
\vspace{-2mm}
Standard 3DGS pipelines rely heavily on COLMAP~\cite{colmap} for initialization. However, under severe motion blur, the high-frequency features required for classical Structure-from-Motion (SfM) are obliterated, causing downstream deblurring methods (e.g., ExBluRF~\cite{lee2023exblurf}, Deblur-GS~\cite{chen2024deblur}, and BAD-Gaussians~\cite{zhao2024badgaussians}) to fail entirely. PRISM3D overcomes this catastrophic failure by explicitly coupling robust deep tracking with probabilistic geometric refinement.

\vspace{-4mm}
\subsubsection{Robust Deep Initialization and Probabilistic Refinement}
Accurate initialization is critical yet challenging when classical feature-based methods collapse. We resolve this by coupling deep dense tracking for global registration with a probabilistic formulation for geometric densification.

\textbf{Robust Initialization via VGGSfM:} 
Unlike incremental SfM pipelines, VGGSfM~\cite{wang2024vggsfm} estimates global camera poses end-to-end. By leveraging a Transformer-based architecture for differentiable point tracking and bundle adjustment via the Theseus solver, it extracts reliable tracks even from severely degraded textures. As evaluated in Section~\ref{sec:robustness_blur_levels}, this yields valid initializations at extreme blur levels where traditional pipelines fail to establish correspondences.

\textbf{Probabilistic Refinement via 3DGS-MCMC:} 
While VGGSfM recovers global topology, the resulting point cloud is inherently sparse. Standard 3DGS heuristic cloning often fails to bridge these structural gaps. We therefore adopt a probabilistic formulation building upon 3DGS-MCMC~\cite{MCMC}, treating Gaussians as samples from an underlying scene distribution. Updated via Stochastic Gradient Langevin Dynamics (SGLD), these primitives dynamically relocate to high-likelihood regions, effectively populating the sparse deep priors without relying on rigid density control rules.

\textbf{Synergistic Joint Optimization:} 
The integration of deep tracking and MCMC creates a synergistic differentiable pipeline. While 3DGS-MCMC adaptively densifies the VGGSfM output, our joint optimization allows gradients to flow back through the continuous Bézier trajectories. This co-refinement ensures that both camera motion and scene geometry remain strictly consistent with the physical and photometric observations of the blurry input.

\vspace{-4mm}
\subsubsection{Physical Motion Blur Image Formation Model}
\label{sec:blur_model_appendix}
Motion blur arises from the accumulation of photons during exposure, expressed mathematically as a continuous integration over virtual latent sharp images $\mathbf{C}_t(\mathbf{u})$ (left), which we approximate using $n$ discrete samples (right):

%\vspace{-4mm}
\begin{minipage}{.48\linewidth}
\begin{equation}\label{eq_continuous_blur_im_formation}
\mathbf{B}(\mathbf{u}) = \phi \int_{0}^{\tau} \mathbf{C}_t(\mathbf{u}) \, dt
\end{equation}
\end{minipage}%
\hfill
\begin{minipage}{.48\linewidth}
\begin{equation}\label{eq_blur_im_formation}
\mathbf{B}(\mathbf{u}) \approx \frac{1}{n} \sum_{i=0}^{n-1} \mathbf{C}_i(\mathbf{u})
\end{equation}
\end{minipage}
\vspace{1mm}

Here, $\mathbf{B}(\mathbf{u})$ is the captured blurry image, $\phi$ is a normalization factor, and $\tau$ is the exposure time. Crucially, this discrete formulation ensures $\mathbf{B}(\mathbf{u})$ remains differentiable with respect to each virtual sharp image $\mathbf{C}_i(\mathbf{u})$, enabling joint optimization.

\vspace{-4mm}
\subsubsection{Continuous Camera Trajectory Modeling}
We parameterize camera poses in the $SE(3)$ Lie group. Because linear or cubic splines~\cite{zhao2024badgaussians} are inadequate for modeling the complex physical momentum of severe camera shake, we adopt a continuous differentiable Bézier curve interpolation~\cite{chen2024deblur}. For a curve of degree $M$ with control points $T_j$, the interpolated pose at time $t$ is:
\begin{equation}
T_t = \prod_{j=0}^{M} \exp \left( \binom{M}{j} (1 - u)^{M-j} u^j \cdot \log(T_j) \right)
\end{equation}
where $u = t/\tau \in [0,1]$. This mathematically ensures smooth, physics-consistent trajectory estimation for accurate deblurring.

\vspace{-4mm}
\subsubsection{Loss Functions and Joint Optimization}
\label{loss_function}
Following \cite{zhao2024badgaussians}, we jointly estimate the 3DGS parameters $\mathbf{\Theta}$ (mean $\boldsymbol{\mu}$, covariance $\mathbf{\Sigma}$, opacity $o$, color $\mathbf{c}$) and the Bézier control points $\mathbf{T}$ by minimizing a combined photometric loss against the captured blurry image $\mathbf{B}^{gt}_k(\mathbf{u})$:
\begin{equation}
\mathcal{L} = (1-\lambda) \mathcal{L}_1 + \lambda \mathcal{L}_{\text{D-SSIM}}
\end{equation}

To optimize $\mathbf{\Theta}$ and $\mathbf{T}$, we compute analytical Jacobians to ensure proper gradient flow~\cite{zhao2024badgaussians} (for clarity, explicit dependence on the pixel coordinate $\mathbf{u}$ is omitted):

\vspace{-2mm}
\noindent\begin{minipage}{.45\linewidth}
\small
\begin{equation}
\frac{\partial \mathcal{L}}{\partial \mathbf{\Theta}} = \sum_{k=0}^{K-1} \frac{\partial \mathcal{L}}{\partial \mathbf{B}_k} \cdot \frac{1}{n} \sum_{i=0}^{n-1} \frac{\partial \mathbf{B}_k}{\partial \mathbf{C}_i} \frac{\partial \mathbf{C}_i}{\partial \mathbf{\Theta}}
\label{eq:grad_theta}
\end{equation}
\end{minipage}%
\hfill
\begin{minipage}{.53\linewidth}
\small
\begin{equation}
\frac{\partial \mathcal{L}}{\partial \mathbf{T}} = \sum_{k=0}^{K-1} \frac{\partial \mathcal{L}}{\partial \mathbf{B}_k} \cdot \frac{1}{n} \sum_{i=0}^{n-1} \frac{\partial \mathbf{B}_k}{\partial \mathbf{C}_i} \frac{\partial \mathbf{C}_i}{\partial \mathbf{\Theta}} \frac{\partial \mathbf{\Theta}}{\partial \mathbf{T}}
\label{eq:grad_pose}
\end{equation}
\end{minipage}
\vspace{2mm}

\noindent Here, camera poses are represented as 6D vectors in their $SE(3)$ Lie algebra.
\vspace{-4mm}
\subsection{PRISM3D-E (Event-Assisted Extreme Motion 3D Deblurring)}
\label{sec_4.3}
While our standalone PRISM3D pipeline is highly capable, we introduce \textbf{PRISM3D-E} to achieve even higher reconstruction quality when complementary event data happens to be available. By leveraging these events in severely degenerate scenarios, we synthesize intermediate sharp images exclusively to bootstrap the VGGSfM initialization. Crucially, these event-derived images are never used for photometric supervision; the final probabilistic optimization remains strictly guided by the original blurry inputs and our physical blur model.

\begin{figure}[ht]
   \vspace{-5mm}
    \centering
    \includegraphics[width=0.9\linewidth]{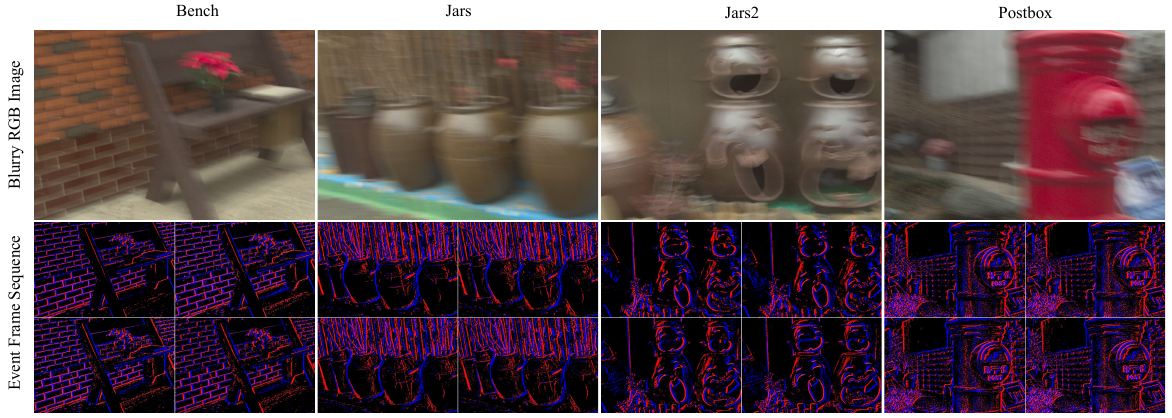}
    \caption{\textbf{The PRISM3D-E Benchmark:} We introduce a synthetic event dataset designed for extreme motion blur. While RGB frames suffer from severe degradation, the complementary event streams preserve fine structural cues.}
    \label{fig:Our_Event_Dataset_Main}
    \vspace{-12mm}
\end{figure}

\subsubsection{Event Generation and Formulation}
\label{sec_3.2}
Event cameras asynchronously trigger an event $\mathbf{e}(x, y, \tau, p)$ when the logarithmic brightness change exceeds a threshold $\Theta$:
\begin{equation}
    p_{x,y,\tau} = 
    \begin{cases}
    -1,&\log(\mathcal{I}_{x,y,\tau}) - \log(\mathcal{I}_{x,y,\tau - \Delta \tau}) < -\Theta\\
    +1,&\log(\mathcal{I}_{x,y,\tau}) - \log(\mathcal{I}_{x,y,\tau - \Delta \tau}) > \Theta
    \end{cases}
\label{eq:6}
\end{equation}
where $p$ is the polarity. To facilitate processing and integration with the EDI model, we group these asynchronous events uniformly into $b$ discrete bins over the exposure time $t_{\text{exp}}$:
\begin{equation}
B_k = \{\mathbf{e}_i(x_i, y_i, \tau_i, p_i)\}_{t_{k-1} < \tau_i \leq t_k}, \quad \text{where } t_k = t_{\text{start}} + \frac{k}{b} t_{\text{exp}}
\label{eq:7}
\end{equation}

\section{Experiments}
\label{sec:experiments}
\vspace{-2mm}
\subsection{Experimental Setup}
\vspace{-2mm}
\paragraph{Datasets:} 
For evaluation, we use the synthetic dataset from ExBluRF~\cite{lee2023exblurf}, comprising eight diverse outdoor scenes with challenging camera motion (20 to 40 blurry training views and 4 to 6 test views per scene). Utilizing the provided latent sharp image sequences, we process the frames via ESIM~\cite{rebecq2018esim} to generate corresponding high-temporal-resolution event streams. This creates our complementary \textbf{PRISM3D-E Benchmark} (Figure~\ref{fig:Our_Event_Dataset_Main}), which we will release publicly to support event-based deblurring research in extreme motion scenarios. The complete event sequences are provided in the supplementary material. 

Additionally, we validate on the real-world E2NeRF dataset~\cite{qi2023e2nerf}, captured using a DAVIS346 color event camera. It features five challenging scenes (\ie, \textit{letter, lego, camera, plant, toys}) with complex textures and irregular hand-held motion (100ms exposure), resulting in severe blur and complex trajectories.

\paragraph{Baseline Methods and Evaluation Metrics:}
We compare PRISM3D against three baseline categories: (1) \textbf{Deep Single-Image Deblurring:} MPRNet~\cite{zamir2021multi} and Restormer~\cite{zamir2022restormer}; (2) \textbf{Event-Based Reconstruction:} EDI~\cite{pan2019bringing}, E2NeRF~\cite{qi2023e2nerf}, and EBADNeRF~\cite{ebadnerf}; and (3) \textbf{Oracle Baselines (Sharp Supervision):} ExBluRF*~\cite{lee2023exblurf}, Deblur-GS*~\cite{chen2024deblur}, and BAD-Gaussians*~\cite{zhao2024badgaussians}. While BAD-Gaussians is theoretically a blind method designed for blurred inputs, its critical dependence on COLMAP initialization renders it an "Oracle" baseline in our extreme blur regime, as classical SfM fails to provide the necessary poses and point clouds from degraded frames. Consequently, ExBluRF*, Deblur-GS*, and BAD-Gaussians* rely on the oracle assumption that accurate poses and point clouds are obtainable from sharp images. Since this is impossible in real-world blind deblurring, there are no direct comparisons with these oracle methods on the real dataset (Figure~\ref{fig:comparisons_real}). We retain event-based methods as valid baselines because EDI-deblurred images enable COLMAP initialization even under severe blur. We evaluate reconstruction quality using \textbf{PSNR}, \textbf{SSIM}, and \textbf{LPIPS}.

\paragraph{Implementation Details:}
Our method is implemented in PyTorch~\cite{paszke2019pytorch} within the \textbf{3DGS-MCMC}~\cite{MCMC} framework using the \textit{gsplat}~\cite{gsplat} pipeline. We optimize Gaussians and camera poses in $SE(3)$ \textbf{Bézier space} (using 9 control points) with the Adam optimizer. Gaussian learning rates follow the original 3DGS~\cite{kerbl20233d}, while the pose learning rate is $1 \times 10^{-3}$. We use 15 virtual camera samples ($n$ in Eq.~\ref{eq_blur_im_formation}) to balance performance and efficiency. For PRISM3D-E, we use 13 event bins for event-based deblurring (EDI). All models are trained on an NVIDIA RTX 4090 GPU using 7k iterations with a data downsampling factor of 2.
\vspace{-4mm}
\subsection{Quantitative Results}
\vspace{-5mm}
\paragraph{Reconstruction Quality:}  
Table~\ref{tab:quan_table} organizes methods into three categories for fair comparison: (1) \textbf{Blind RGB} (w/o Events), (2) \textbf{Event-Assisted} (w/ Events), and (3) \textbf{Oracle/Impractical} (w/ Sharp Supervision). 
%\vspace{-5mm}
\begin{table}[ht]
    \centering
    \caption{\textbf{Quantitative comparisons for sharp novel view synthesis on the PRISM3D-E Benchmark (Synthetic).} 
The table is grouped by input modality: (1) \textbf{Blind RGB} (w/o Events), (2) \textbf{Event-Assisted} (w/ Events), and (3) \textbf{Oracle Baselines} (w/ Sharp Supervision). Group 3 methods (ExBluRF*, BAD-Gaussians*, Deblur-GS*) are included only for reference as they rely on impractical sharp-image initialization. Best and second-best results within ranked groups are highlighted in \colorbox{green!25}{green} and \colorbox{red!25}{orange}. \emph{Note: For brevity, we display three representative scenes matching our qualitative comparisons (Figure~\ref{fig:comparisons_synth}); however, the \textbf{Average} row is computed over the entire 8-scene dataset. The full per-scene breakdown is provided in the supplementary material.}}
    \label{tab:quan_table}
    \renewcommand{\arraystretch}{1.35}
    \resizebox{0.95\linewidth}{!}{
        \begin{tabular}{c||c||ccc|cccc|ccc}
            \toprule
            & & \multicolumn{3}{c|}{\textbf{Blind RGB (w/o Events)}} & \multicolumn{4}{c|}{\textbf{Event-Assisted (w/ Events)}} & \multicolumn{3}{c}{\textbf{Oracle Baselines(*)}} \\
            \textbf{Scene} & \textbf{Metric} & \textbf{MPRNet} & \textbf{Restormer} & \textbf{PRISM3D} & \textbf{EDI+3DGS} & \textbf{E2NeRF} & \textbf{EBAD-NeRF} & \textbf{PRISM3D-E} & \textbf{ExBluRF*} & \textbf{BAD-Gaussians*} & \textbf{Deblur-GS*} \\
            \hhline{=::=::===::====::===}
            \multirow{3}{*}{\textbf{Bench}} 
                & PSNR$\uparrow$ & 25.35 & \colorbox{red!25}{26.39} & \colorbox{green!25}{29.86} & \colorbox{red!25}{28.95} & {25.41} & {28.15} & \colorbox{green!25}{33.55} & 31.93 & 32.54 & 33.26 \\
                & SSIM$\uparrow$ & 0.678 & \colorbox{red!25}{0.720} & \colorbox{green!25}{0.841} &  \colorbox{red!25}{0.865} & {0.708} &{0.822} & \colorbox{green!25}{0.924} & 0.877 & 0.901 & 0.929 \\
                & LPIPS$\downarrow$ & 0.425 & \colorbox{red!25}{0.356} & \colorbox{green!25}{0.118} & 0.201 & {0.438} & \colorbox{red!25}{0.172} & \colorbox{green!25}{0.063} & 0.111 & 0.046 & 0.116 \\
            \hhline{-||-||----------}
            \multirow{3}{*}{\textbf{Camellia}} 
                & PSNR$\uparrow$ & 24.84 & \colorbox{red!25}{25.14} & \colorbox{green!25}{28.56} & 22.46 & \colorbox{red!25}{28.07} & 24.33 & \colorbox{green!25}{29.47} & 28.02 & 28.83 & 29.14 \\
                & SSIM$\uparrow$ & 0.669 & \colorbox{red!25}{0.690} & \colorbox{green!25}{0.821} & \colorbox{red!25}{0.762} & {0.721} & {0.743} & \colorbox{green!25}{0.873} & 0.715 & 0.815 & 0.874 \\
                & LPIPS$\downarrow$ & 0.395 & \colorbox{red!25}{0.351} & \colorbox{green!25}{0.129} & 0.271 & {0.329} & \colorbox{red!25}{0.192} & \colorbox{green!25}{0.108} & 0.313 & 0.099 & 0.158 \\
            \hhline{-||-||----------}
            \multirow{3}{*}{\textbf{Stone Lantern}} 
                & PSNR$\uparrow$ & 24.97 & \colorbox{red!25}{26.68} & \colorbox{green!25}{28.29} & 26.48 & \colorbox{green!25}{30.47} & 26.29 & \colorbox{red!25}{29.43} & 28.24 & 28.29 & 28.59 \\
                & SSIM$\uparrow$ & 0.785 & \colorbox{red!25}{0.831} & \colorbox{green!25}{0.849} & 0.825 & \colorbox{red!25}{0.836} & 0.802 & \colorbox{green!25}{0.894} & 0.765 & 0.843 & 0.863 \\
                & LPIPS$\downarrow$ & 0.342 & \colorbox{red!25}{0.280} & \colorbox{green!25}{0.195} & 0.270 & {0.324} & \colorbox{red!25}{0.264} & \colorbox{green!25}{0.152} & 0.236 & 0.143 & 0.236 \\
            \hhline{=::=::===::====::===}
            \multirow{3}{*}{\textbf{Average (All 8)}} & PSNR$\uparrow$ & 26.19 & \colorbox{red!25}{26.83} & \colorbox{green!25}{29.49} & 27.55 & \colorbox{red!25}{29.35} & 28.36 & \colorbox{green!25}{31.95} & 31.15 & 30.95 & 31.43 \\
                & SSIM$\uparrow$ & 0.733 & \colorbox{red!25}{0.756} & \colorbox{green!25}{0.840} & 0.830 & {0.771} & \colorbox{red!25}{0.833} & \colorbox{green!25}{0.908} & 0.834 & 0.863 & 0.895 \\
                & LPIPS$\downarrow$ & 0.368 & \colorbox{red!25}{0.324} & \colorbox{green!25}{0.153} & 0.232 & {0.352} & \colorbox{red!25}{0.183} & \colorbox{green!25}{0.097} & 0.162 & 0.089 & 0.178 \\
            \bottomrule
        \end{tabular}
    }
    \vspace{-4mm}
\end{table}

Operating solely on motion-blurred images, \textbf{PRISM3D} outperforms all competing blind methods and several event-based approaches, proving the efficacy of probabilistic MCMC refinement for sparse initialization. When event data is available, \textbf{PRISM3D-E} achieves an average \textbf{2.5 dB PSNR gain} over one of the state-of-the-art event-based methods, E2NeRF. Notably, \textbf{PRISM3D-E delivers 0.5 - 0.8 dB PSNR improvement over sharp-supervised oracle baselines (ExBluRF*, BAD-Gaussians*, Deblur-GS*)} despite their privileged sharp image access. This highlights the immense strength of integrating event-based priors, VGGSfM initialization, and 3DGS-MCMC joint optimization. 

\begin{table}[!ht] 
    \vspace{-4mm}
    \centering
    \begin{minipage}[c]{0.52\textwidth} 
        \scriptsize
        \renewcommand{\arraystretch}{1.1}
        \caption{Performance comparison between PRISM3D and BAD-Gaussians (SOTA) across various blur levels.}
        \label{tab:metrics_prism3d_bad_gaussians}
        \resizebox{\linewidth}{!}{%
        \begin{tabular}{@{}l ccc ccc@{}}
        \toprule
        \multirow{2}{*}{\textbf{Blur Level}} & \multicolumn{3}{c}{\textbf{PRISM3D (Ours)}} & \multicolumn{3}{c}{\textbf{BAD-Gaussians}} \\
        \cmidrule(lr){2-4} \cmidrule(l){5-7}
        & \textbf{PSNR$\uparrow$} & \textbf{SSIM$\uparrow$} & \textbf{LPIPS$\downarrow$} & \textbf{PSNR$\uparrow$} & \textbf{SSIM$\uparrow$} & \textbf{LPIPS$\downarrow$} \\
        \midrule
        Sharp & \textbf{41.42} & \textbf{0.975} & \textbf{0.016} & 41.02 & 0.974 & 0.017 \\
        3     & \textbf{37.35} & \textbf{0.949} & 0.038 & 37.14 & 0.946 & \textbf{0.036} \\
        5     & \textbf{36.70} & \textbf{0.942} & \textbf{0.032} & 33.72 & 0.901 & 0.069 \\
        7     & \textbf{35.33} & \textbf{0.928} & \textbf{0.037} & 30.99 & 0.842 & 0.111 \\
        9     & \textbf{35.09} & \textbf{0.924} & \textbf{0.035} & \textcolor{red}{x} & \textcolor{red}{x} & \textcolor{red}{x} \\
        11    & \textbf{32.22} & \textbf{0.897} & \textbf{0.093} & \textcolor{red}{x} & \textcolor{red}{x} & \textcolor{red}{x} \\
        \bottomrule
        \end{tabular}%
        }
    \end{minipage}
    \hfill
    \begin{minipage}[c]{0.44\textwidth} 
        \paragraph{Robustness vs. BAD-Gaussians:}
        \scriptsize
        {As shown in Table~\ref{tab:metrics_prism3d_bad_gaussians}, while BAD-Gaussians applies COLMAP directly to blurred inputs, their approach is primarily evaluated on mild motion blur. Relying on classical SfM becomes a bottleneck under extreme degradation. Because severe blur smears the high-frequency details (e.g., sharp corners and edges) required for reliable feature matching, COLMAP cannot establish correspondences and fails completely beyond 7 frames. In contrast, PRISM3D's deep tracking initialization allows it to maintain robust reconstruction up to a severe 11-frame blur.}
    \end{minipage}
    \vspace{-4mm}
\end{table}

\vspace{-2mm}
\paragraph{Why PRISM3D-E Outperforms Oracle Baselines:}
Our framework couples robust initialization with physics-consistent trajectory modeling. \textbf{MCMC refinement} is more resilient to pose errors than the deterministic heuristics used in BAD-Gaussians. Furthermore, our continuous \textbf{Bézier trajectory model} (Figure~\ref{fig:trajectory_plots}) better approximates real-world motion blur than linear or cubic splines. EDI-deblurred images also allow VGGSfM to estimate highly accurate camera poses (\textbf{mean APE of 0.0862}), which is crucial for reliable 3D reconstruction. Finally, unlike ExBluRF, whose voxel-based representation struggles with high-frequency content, our bundle-adjusted Gaussian representation maintains superior geometric fidelity, allowing PRISM3D-E to surpass oracle baselines without sharp supervision.

\paragraph{Training Time and GPU Memory Consumption:}

\begin{wraptable}{r}{0.5\textwidth}
    \vspace{-10mm} 
    \centering
    \caption{\textbf{System Efficiency (Real Dataset).} Average training time and GPU memory consumption.}
    \label{tab:system_efficiency}
    \renewcommand{\arraystretch}{1.2}
    \resizebox{\linewidth}{!}{
        \begin{tabular}{l | c c}
            \toprule
            \textbf{Method} & \textbf{Time (hh:mm:ss)$\downarrow$} & \textbf{Memory (MiB)$\downarrow$} \\
            \midrule
            EBAD-NeRF & 06:44:00 & 14,836 \\
            E2NeRF & 14:58:00 & 15,532 \\
            \rowcolor{gray!15} \textbf{PRISM3D (Ours)} & \textbf{00:07:13} & \textbf{1,584} \\
            \bottomrule
        \end{tabular}
    }
    \vspace{-0.15in} 
\end{wraptable}

Evaluated on the real dataset (Table~\ref{tab:system_efficiency}), PRISM3D completes optimization in just \textbf{7 minutes} per scene. We exclude explicit baselines (e.g., BAD-Gaussians) here, as their COLMAP dependency causes initialization failure on these severely degraded sequences. Conversely, surviving event-based implicit baselines (EBAD-NeRF, E2NeRF) require 7--15 hours and $\sim$15 GiB of GPU memory. In contrast, PRISM3D consumes only \textbf{$\sim$1.55 GiB}, confirming the extreme efficiency and scalability of our representation.

\vspace{-4mm}
\subsection{Qualitative Results}  
\vspace{-2mm}
Qualitative comparisons on synthetic and real datasets are presented in Figures~\ref{fig:comparisons_synth} and \ref{fig:comparisons_real}. PRISM3D and PRISM3D-E consistently produce sharper reconstructions with fewer artifacts than existing approaches. In synthetic scenes like  \textit{Bench}, \textit{Camellia} and \textit{Stone Lantern}, competing methods over-smooth fine structures or introduce distortions (\textbf{red arrows}, Figure~\ref{fig:comparisons_synth}), whereas PRISM3D and its multi-modal extension, PRISM3D-E faithfully restores object fine details. 
\begin{figure}[!ht]
    \centering
    \vspace{-6mm}
    \includegraphics[width=0.99\linewidth]{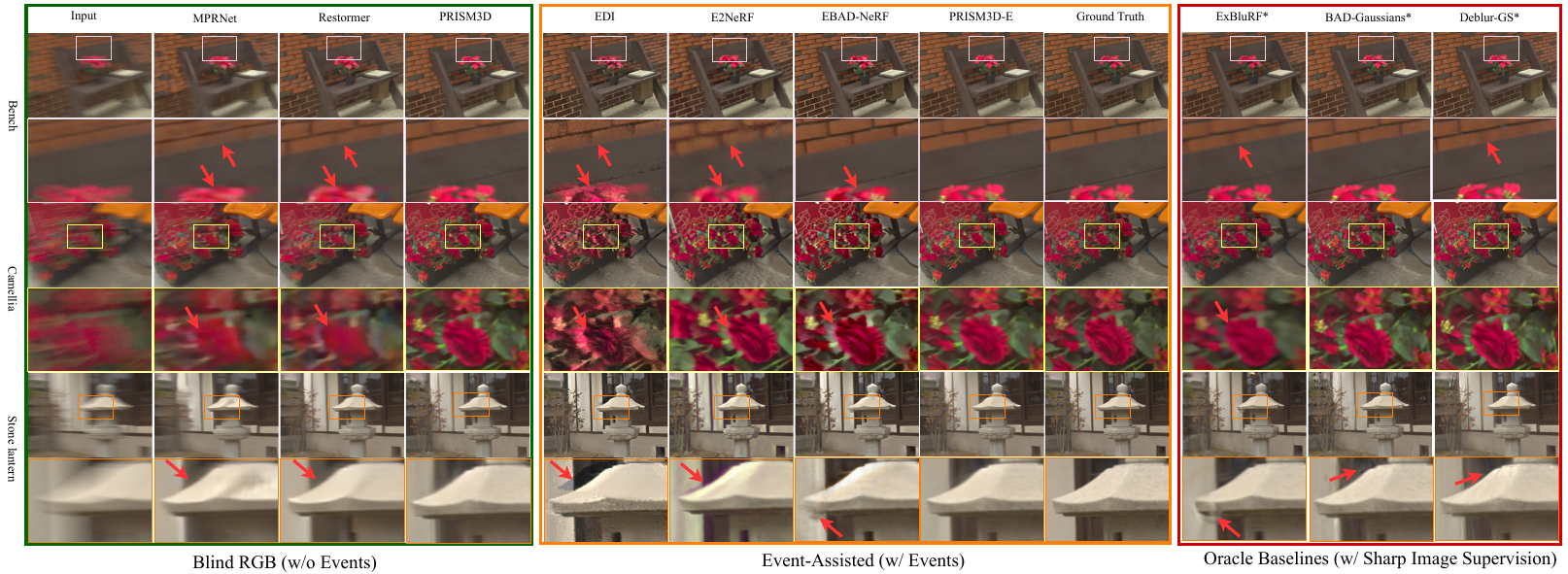}
 \caption{\textbf{Results on the Synthetic Dataset:} \textbf{PRISM3D} effectively resolves extreme motion blur, preserving fine structural details and minimizing artifacts compared to existing methods. Note that ExBluRF*, BAD-Gaussians*, and Deblur-GS* utilize an oracle initialization (sharp images for poses and point clouds), whereas our method operates blindly.}
    \label{fig:comparisons_synth}
    \vspace{-6mm}
\end{figure}

\begin{figure}[!ht]
    \centering
    %\vspace{-4mm}
    \includegraphics[width=0.95\linewidth]{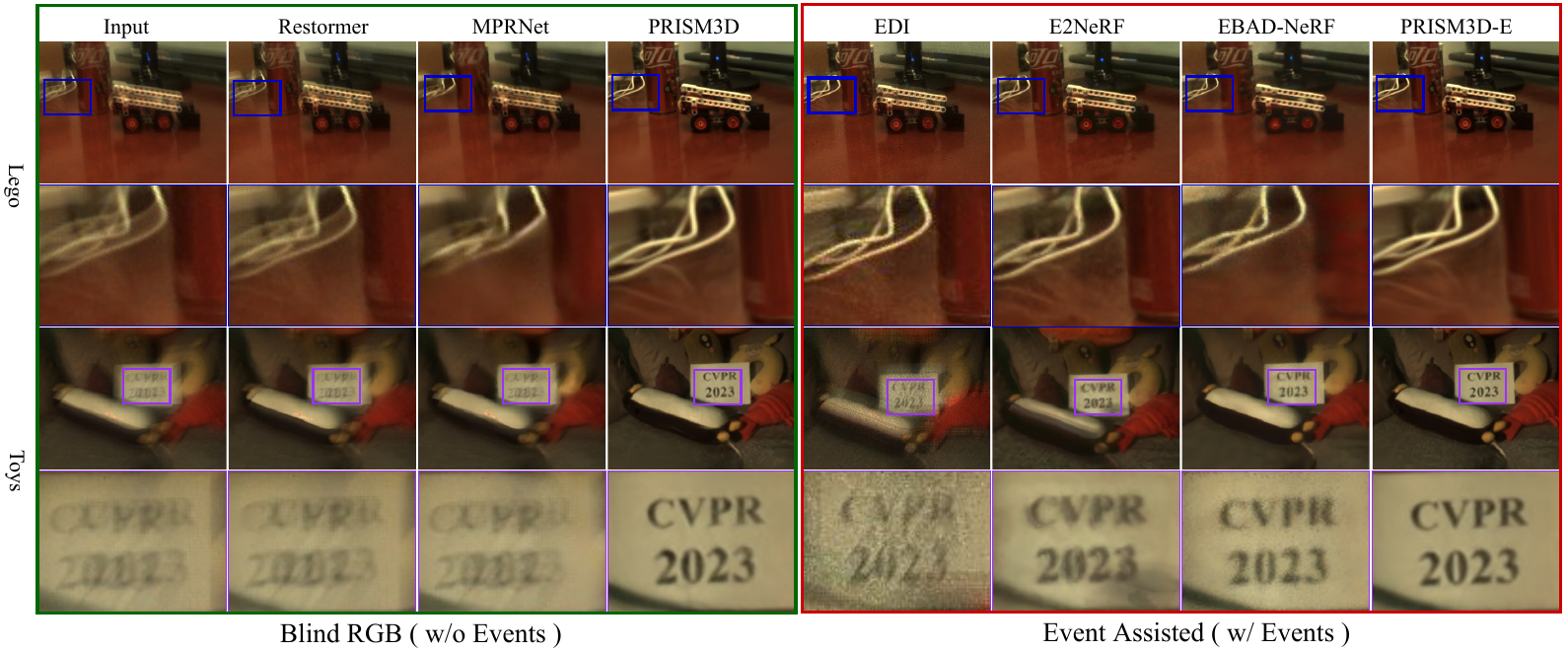}
 \caption{\textbf{Results on the Real Dataset:} PRISM3D reconstructs sharp and high-quality images from severely motion-blurred real-world inputs. In contrast, existing methods struggle with artifacts, noise, loss of fine details, and text degradation. Our framework effectively restores textures and structural consistency, as evident in the insets.}
    \label{fig:comparisons_real}
    \vspace{-6mm}
\end{figure}
In real-world scenarios (Figure~\ref{fig:comparisons_real}), our method achieves superior clarity in high-frequency regions, recovering legible text on the \textbf{\textit{CVPR 2023} poster} and sharp specular highlights. While prior event-based methods (E2NeRF, EBAD-NeRF) suffer from color distortions and residual blur, PRISM3D and its multi-modal extension preserve accurate color distributions and fine-grained textures without ghosting artifacts.

\vspace{-4mm}
\subsection{Robustness of Modules across Various Blur Levels}
\label{sec:robustness_blur_levels}

To comprehensively evaluate our framework, we assess the robustness of each module across progressive motion blur levels before conducting targeted analyses in extreme blur scenarios to isolate individual contributions.

\begin{figure}[!ht] % Changed from figure* to figure[t]
    \vspace{-6mm} % Pulls the image up slightly
    \centering
    \includegraphics[width=\linewidth]{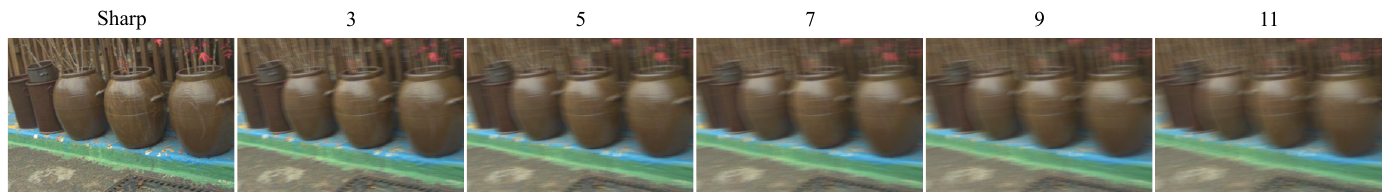}
    %\vspace{-3mm} % Pulls the caption closer to the image
    \caption{\textbf{Synthetic Motion Blur Levels:} Images generated by averaging 1 (sharp), 3, 5, 7, 9, and 11 consecutive frames from an 11-image burst at each viewpoint, illustrating the increasing severity of motion blur used for evaluation.}
    \label{fig:Blur_Levels}
    \vspace{-8mm} % Pulls the text below closer to the figure
\end{figure}

\vspace{-4mm}
\subsubsection{VGGSfM Robustness:}
To evaluate VGGSfM's robustness, we simulated increasing blur severity by averaging 1, 3, 5, 7, 9, and 11 consecutive frames from sharp burst sequences (Figure~\ref{fig:Blur_Levels}). While COLMAP fails to register images beyond moderate blur, VGGSfM consistently extracts stable poses and dense point clouds even under extreme degradation (Table~\ref{tab:colmap_vs_vggsfm}). Translating this to quantitative accuracy, VGGSfM maintains highly reliable camera poses compared to ground truth (Table~\ref{tab:vggsfm_pose_error_stats}), demonstrating its superiority.

\begin{table}[!ht] 
    \vspace{-4mm}
    \centering
    \scriptsize

    % First: SfM comparison table
    \begin{minipage}[b]{0.48\textwidth}
        \renewcommand{\arraystretch}{1.1} 
        \centering
        \caption{SfM comparison: Performance across various blur levels.}
        \label{tab:colmap_vs_vggsfm}
        \begin{tabular}{l|cc|cc}
        \toprule
        \multirow{2}{*}{Blur} & \multicolumn{2}{c|}{VGGSfM} & \multicolumn{2}{c}{COLMAP} \\
        \cmidrule(lr){2-3} \cmidrule(lr){4-5}
        & \#Pts & \#Imgs & \#Pts & \#Imgs \\
        \midrule
        Sharp & 23498 & 20 & 4987 & 20 \\
        3     & 22873 & 20 & 1257 & 20 \\
        5     & 21660 & 20 & 606  & 20 \\
        7     & 18518 & 20 & 246  & 15 \\
        9     & 21351 & 20 & \textcolor{red}{x}   & \textcolor{red}{x} \\
        11    & 20283 & 20 & \textcolor{red}{x}   & \textcolor{red}{x} \\
        \bottomrule
        \end{tabular}
    \end{minipage}\hfill
    % Second: Pose error stats table
    \begin{minipage}[b]{0.48\textwidth}
        \renewcommand{\arraystretch}{1.38} % Compensates for the missing 'Sharp' row
        \centering
        \caption{VGGSfM Pose translation error statistics (in meters) with respect to ground truth.}
        \label{tab:vggsfm_pose_error_stats}
        \begin{tabular}{lccccc}
        \toprule
        Blur & RMSE $\downarrow$ & Mean $\downarrow$ & Med. $\downarrow$ & Std $\downarrow$ & Max $\downarrow$ \\
        \midrule
        3   & 0.18 & 0.16 & 0.13 & 0.09 & 0.35 \\
        5   & 0.32 & 0.27 & 0.20 & 0.18 & 0.65 \\
        7   & 0.40 & 0.36 & 0.29 & 0.17 & 0.75 \\
        9   & 0.61 & 0.49 & 0.36 & 0.36 & 1.43 \\
        11  & 0.72 & 0.52 & 0.35 & 0.50 & 2.24 \\
        \bottomrule
        \end{tabular}
    \end{minipage}
\end{table}
\vspace{-4mm}
\subsubsection{MCMC Robustness:}

\begin{figure}[!ht]
    \vspace{-4mm}
    \centering
    \begin{minipage}[t]{0.48\textwidth}
        \centering
        \includegraphics[width=\linewidth]{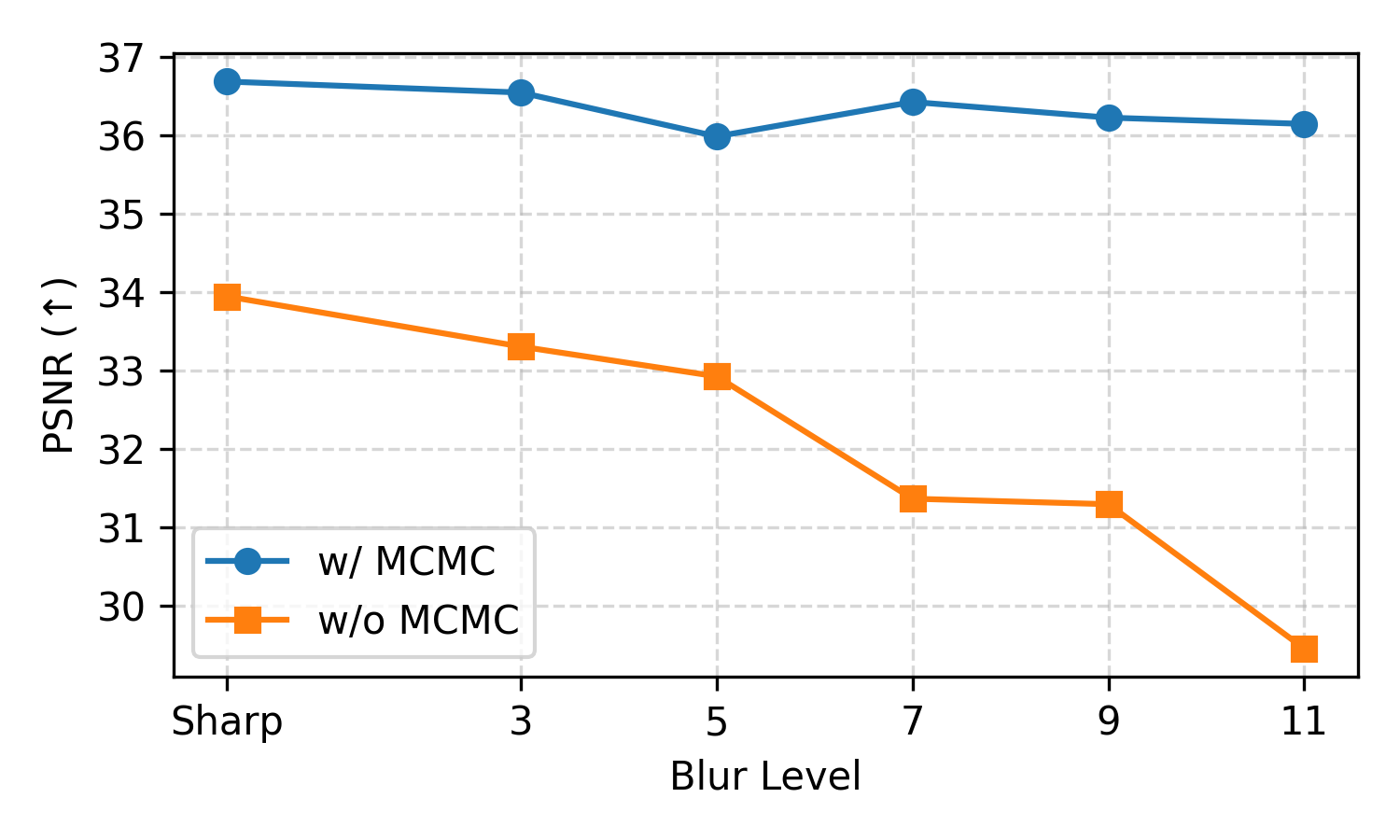}
        \vspace{-6mm}
        \caption{\textbf{MCMC robustness to various blur-corrupted point cloud initializations:}   PSNR comparison of 3DGS-MCMC with 3DGS across various blur point cloud initializations obtained from VGGSfM.}
        \label{fig:psnr_plot_blur_point_clouds_3dgs_mcmc}
    \end{minipage}
    \hfill
    \begin{minipage}[t]{0.48\textwidth}
        \centering
        \includegraphics[width=\linewidth]{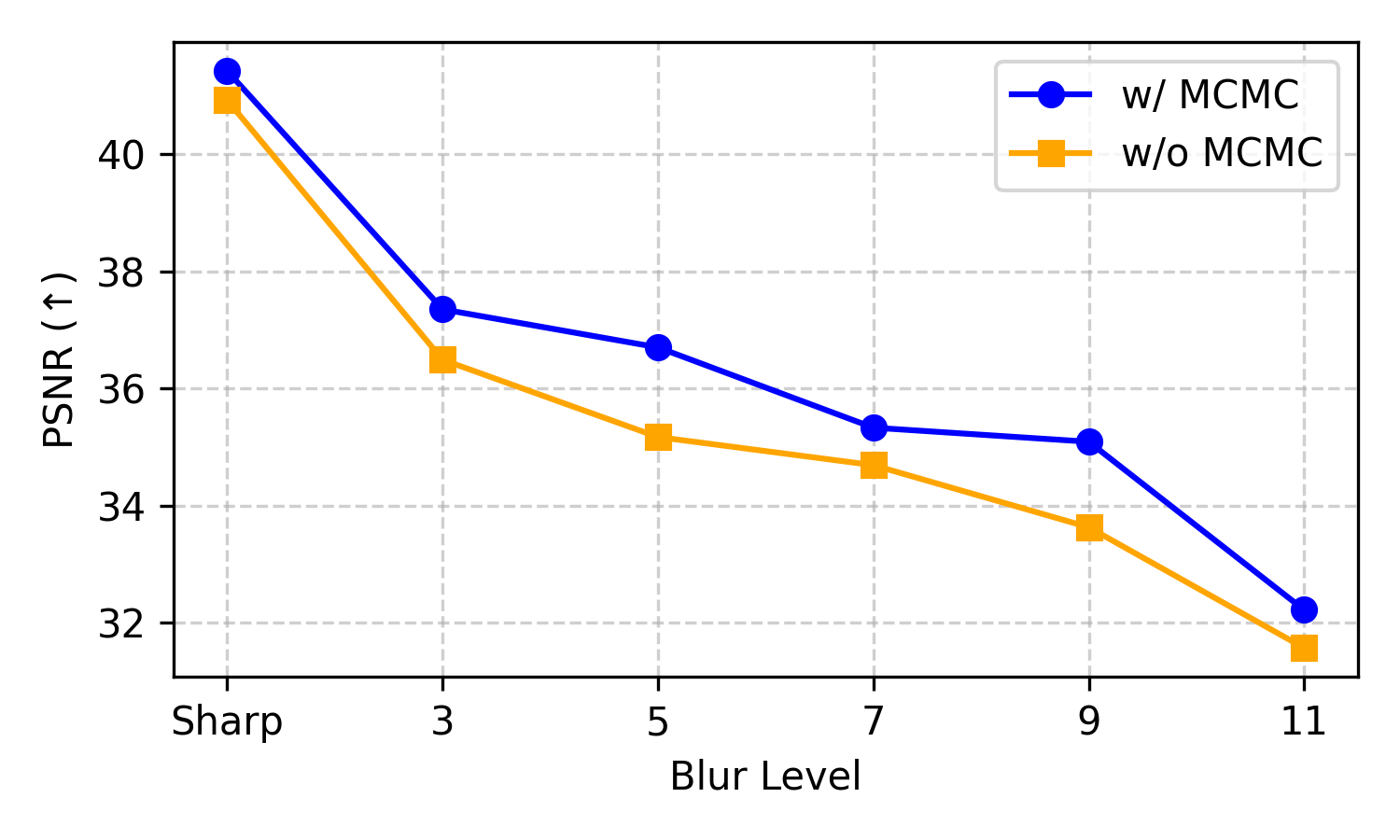}
        \vspace{-6mm}
        \caption{\textbf{MCMC robustness to various blur levels:} PSNR comparison of PRISM3D for deblurring with and without MCMC across various blur levels, SfM initialization are obtained from VGGSfM.}
        \label{fig:pnsr_blur_leves_3dgs_mcmc}
    \end{minipage}
    \vspace{-6mm}
\end{figure}

We conducted two experiments to validate 3DGS-MCMC. First, we evaluated its resilience to noisy blur point cloud initializations by feeding it blur-degraded point clouds from VGGSfM (e.g., blur level 3, blur level 5), while giving the poses estimated from ground truth sharp images. As shown in Figure~\ref{fig:psnr_plot_blur_point_clouds_3dgs_mcmc}, standard 3DGS suffers significant performance drops as blur increases, whereas 3DGS-MCMC maintains stable reconstruction quality, proving the robustness of its probabilistic sampling framework towards varying blur level point clouds. 

Second, we assessed end-to-end deblurring performance within PRISM3D across varying blur levels. As shown in Figure~\ref{fig:pnsr_blur_leves_3dgs_mcmc}, PRISM3D equipped with MCMC consistently achieves superior PSNR. Furthermore, it successfully suppresses the noticeable artifacts and structural inconsistencies present in the non-MCMC variant (see supplementary material for visual comparisons), highlighting its practical value for reliable scene recovery.

\begin{comment}
\paragraph{Comparative Robustness vs. BAD-Gaussians:}
\textcolor{red}{To evaluate downstream practical applicability, we systematically compared PRISM3D against BAD-Gaussians across increasing blur severities. We provided BAD-Gaussians with COLMAP initializations and PRISM3D with VGGSfM initializations. As detailed in the Supplementary Material, PRISM3D consistently outperforms BAD-Gaussians at all comparable blur levels (e.g., maintaining 35.33 dB vs. 30.99 dB at blur level 7). Notably, as blur severity increases, BAD-Gaussians' performance drops sharply and becomes completely inapplicable beyond level 7 due to COLMAP's failure. Conversely, PRISM3D remains robust up to extreme 11-frame blur, highlighting its capacity to handle severe degradation where traditional pipelines collapse.}
\end{comment}

\vspace{-5mm}
\subsection{Virtual Camera Count \& Trajectory Representation}
\begin{figure*}[!ht]
    \centering
    \vspace{-4mm}
    \includegraphics[width=0.8\linewidth]{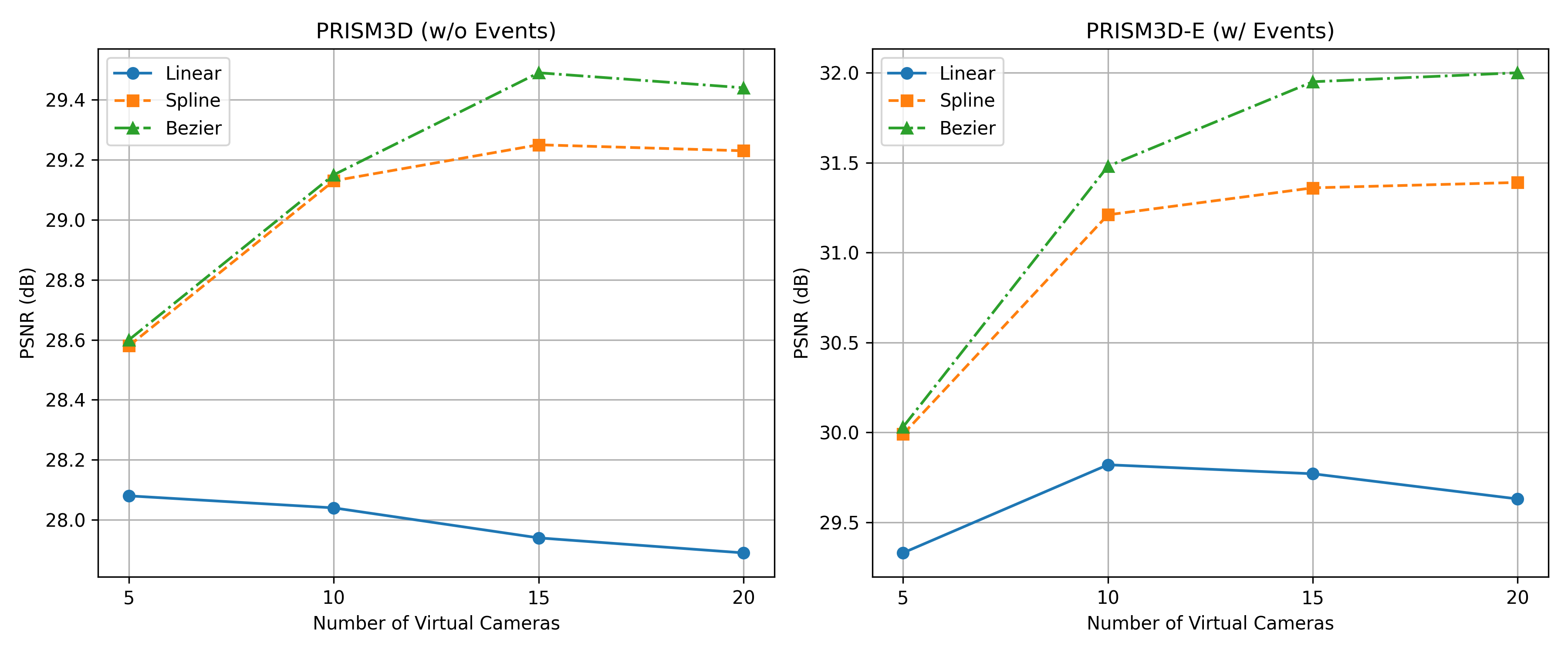}
    \vspace{-4mm}
    \caption{\textbf{Impact of trajectory representations and virtual camera count.} We evaluate linear, spline, and Bézier interpolations across varying discrete trajectory steps ($n$). \textbf{Left:} Standalone PRISM3D. \textbf{Right:} Multi-modal PRISM3D-E. In both regimes, reconstruction quality optimizes at $n=15$. Linear trajectories fail to capture complex intra-exposure motion, while our continuous $SE(3)$ Bézier formulation consistently outperforms cubic splines, proving its necessity for physics-consistent deblurring.}
    \label{fig:trajectory_plots}
    \vspace{-8mm}
\end{figure*}

\vspace{-3mm}
As illustrated in Fig.~\ref{fig:trajectory_plots}, we analyze the number of interpolated virtual camera poses ($n \in \{5,10,15,20\}$) alongside various trajectory models. Across all configurations, increasing $n$ from 5 to 15 yields substantial improvements, beyond which performance saturates. We thus adopt $n=15$ to balance photometric accuracy with computational efficiency. 

Comparing trajectory models, naive linear interpolation consistently underperforms, confirming the non-linear nature of real-world motion blur. While cubic B-splines provide a strong baseline, our continuous $SE(3)$ Bézier formulation consistently achieves the highest performance. This superiority is particularly pronounced in the multi-modal \textbf{PRISM3D-E} setting, where the Bézier trajectory outperforms splines by 0.59 dB in PSNR at $n=15$. This validates that modeling camera momentum via continuous Bézier curves is a fundamental requirement for maximizing geometric recovery across sensor modalities.

%\vspace{-4mm}

\vspace{-4mm}
\subsection{Ablations}
\vspace{-2mm}
\label{sec:ablations}

\subsubsection{Component-wise Analysis for Extreme Motion Blur} 
To quantify the contribution of each module in PRISM3D and PRISM3D-E, we performed ablation studies on the synthetic dataset (Table~\ref{tab:novel_view_synthesis}). Our analysis (Table~\ref{tab:colmap_vs_vggsfm}) shows that classical COLMAP fails entirely under severe motion blur. While the deep feature matcher HLOC successfully runs, it lags behind VGGSfM by an average of 1.45 dB in PSNR. Furthermore, even recent 3D foundation models like VGGT-X struggle in this extreme regime, underperforming our initialization by 1.67 dB. This firmly highlights VGGSfM's superior capability in extracting robust global topology from severe degradation. The inclusion of MCMC-based probabilistic refinement contributes an additional 0.47 dB in PSNR. We also validate our physics-consistent blur formation model. Assuming a naive linear camera trajectory results in a steep 1.55 dB drop. Our continuous $SE(3)$ Bézier formulation outperforms standard spline interpolation by 0.24 dB in the standalone setting. Crucially, when high-temporal-resolution event priors (EDI) are integrated, the necessity of accurate trajectory modeling becomes even more pronounced: our Bézier formulation outperforms linear and spline interpolations by 2.18 dB and 0.59 dB, respectively. Overall, integrating these event-based priors provides the most significant boost, yielding a 2.46 dB PSNR improvement over the standalone PRISM3D. %Ultimately, the full PRISM3D-E pipeline achieves the highest overall performance, producing the sharpest and most faithful reconstructions (Figure~\ref{fig:ablations_real}).

\paragraph{Why Deep SfM over 3D Foundation Models?}
{As highlighted by recent studies on 3D Foundation Models (3DFMs) like VGGT~\cite{vggt}, scaling feed-forward models to dense views yields imperfect outputs with higher noise levels~\cite{vggt-x}. For initialization-sensitive representations like 3DGS, this noise undermines the learning of 3D primitives and traps the highly non-convex optimization process in local minima, leading to significant degradation in rendering quality~\cite{vggt-x}. Our empirical analysis directly confirms this fundamental limitation under extreme blur degradation. As shown in Table~\ref{tab:novel_view_synthesis}, substituting the robust tracking initialization of PRISM3D with VGGT-X causes a severe drop in downstream deblurring quality (from 29.49 dB down to 27.82 dB). This 1.67 dB performance penalty reinforces that the feed-forward geometry of 3DFMs is currently insufficient for blind deblurring. Conversely, because VGGSfM is a dedicated deep SfM pipeline, its intrinsic bundle adjustment and strict multi-view constraints successfully extract the clean, geometrically consistent topological skeleton strictly necessary to guide our probabilistic densification.}
%\vspace{-4mm}
\begin{table}[!ht]
    \vspace{-6mm}
    \centering
    \caption{\textbf{Ablation study.} Average metrics across all 8 synthetic scenes. The inclusion of each module (VGGSfM, MCMC, EDI) and the use of Bézier trajectories yield synergistic improvements.}
    \label{tab:novel_view_synthesis}
    
    \scriptsize % <-- CHANGED: Stepped down from \footnotesize to \scriptsize
    \renewcommand{\arraystretch}{0.3} % (Note: 0.3 is extremely tight, watch for text overlap in the PDF!)
    \setlength{\tabcolsep}{2.5pt} % <-- CHANGED: Reduced column padding from 4pt to 2.5pt
    \setlength{\fboxsep}{0pt}

    % Uncomment the \resizebox line below (and the closing brace at the end) if you need extreme scaling
    % \resizebox{0.95\columnwidth}{!}{
    \begin{tabular}{@{}l cccc ccc@{}}
        \toprule
        \multirow{2}{*}{\textbf{Variant}} & \multicolumn{4}{c}{\textbf{Configuration}} & \multicolumn{3}{c}{\textbf{Metrics}} \\
        \cmidrule(lr){2-5} \cmidrule(l){6-8}
        & \textbf{SfM} & \textbf{MCMC} & \textbf{Traj.} & \textbf{EDI} & \textbf{PSNR$\uparrow$} & \textbf{SSIM$\uparrow$} & \textbf{LPIPS$\downarrow$} \\
        \midrule

        w/ HLOC  & HLOC   & \checkmark & Bézier & $\times$ & 28.04 & 0.761 & \colorbox{red!25}{0.148} \\
        w/ VGGT-X& VGGT-X & \checkmark & Bézier & $\times$ & 27.82 & 0.749 & 0.181 \\
        w/o MCMC & VGGSfM & $\times$   & Bézier & $\times$ & 29.02 & 0.806 & \colorbox{green!25}{0.133} \\
        w/ Linear& VGGSfM & \checkmark & Linear & $\times$ & 27.94 & 0.790 & 0.190 \\
        w/ Spline& VGGSfM & \checkmark & Spline & $\times$ & \colorbox{red!25}{29.25} & \colorbox{red!25}{0.840} & 0.150 \\

        \rowcolor{gray!15}
        \textbf{PRISM3D} & VGGSfM & \checkmark & Bézier & $\times$ & \colorbox{green!25}{29.49} & \colorbox{green!25}{0.840} & 0.153 \\

        \midrule

        w/ COLMAP& COLMAP & \checkmark & Bézier & \checkmark & 31.28 & 0.886 & 0.139 \\
        w/o MCMC & VGGSfM & $\times$   & Bézier & \checkmark & 31.06 & 0.869 & \colorbox{green!25}{0.083} \\
        w/ Linear& VGGSfM & \checkmark & Linear & \checkmark & 29.77 & 0.860 & 0.150 \\
        w/ Spline& VGGSfM & \checkmark & Spline & \checkmark & \colorbox{red!25}{31.36} & \colorbox{red!25}{0.900} & 0.100 \\

        \rowcolor{gray!15}
        \textbf{PRISM3D-E}& VGGSfM & \checkmark & Bézier & \checkmark & \colorbox{green!25}{31.95} & \colorbox{green!25}{0.908} & \colorbox{red!25}{0.097} \\

        \bottomrule
    \end{tabular}
    % } % <-- Uncomment this closing brace if using \resizebox above

    \vspace{-10mm} % WARNING: If this pulls the text below *into* the table, change it to -6mm or -8mm
\end{table}
\begin{comment}
\begin{figure}[!ht]
   %\vspace{-5mm}
    \centering
    \includegraphics[width=0.9\linewidth]{Figures/GEMS_Real_Ablations.pdf}
 \caption{\textbf{Ablation study on different modules on the Real Dataset:} PRISM3D produces high-quality reconstructions, with MCMC effectively suppressing artifacts. Incorporating event data in PRISM3D-E further refines the results.}
    \label{fig:ablations_real}
    %\vspace{-12mm}
\end{figure}
\end{comment}

\subsubsection{Event Modality vs. View Sparsity Analysis}
To comprehensively evaluate the utility of the event modality beyond initialization, we investigate integrating continuous event supervision directly into our core optimization loop via an explicit event reconstruction loss. As summarized in Table~\ref{tab:event_loss_main}, utilizing event priors during the initialization phase (PRISM3D-E Init. Only) yields a substantial $+2.46$\,dB PSNR gain over the blind RGB baseline. Conversely, appending an explicit event constraint to the downstream optimization loss (PRISM3D-E Init. + Event Loss) provides no additional performance improvement and marginally increases rendering artifacts under normal view densities.

\begin{wraptable}{r}{0.55\linewidth}
\vspace{-16mm} % Compresses top margin; adjust as needed
\centering
\caption{Impact of event utilization during initialization vs. continuous optimization.}
\label{tab:event_loss_main}
\resizebox{\linewidth}{!}{
\begin{tabular}{lccc}
\toprule
Method Configuration & PSNR $\uparrow$ & SSIM $\uparrow$ & LPIPS $\downarrow$ \\
\midrule
PRISM3D (RGB Only) & 29.49 & 0.840 & 0.153 \\
PRISM3D-E (Init. + Event Loss) & 31.86 & \textbf{0.909} & 0.098 \\
PRISM3D-E (Init. Only) (Ours) & \textbf{31.95} & 0.908 & \textbf{0.097} \\
\bottomrule
\end{tabular}
}
\vspace{-6mm} % Compresses bottom margin; adjust as needed
\end{wraptable}

\begin{figure}[!h]
\vspace{-6mm}
\centering
\begin{subfigure}[b]{0.4\linewidth}
\centering
\includegraphics[width=\textwidth]{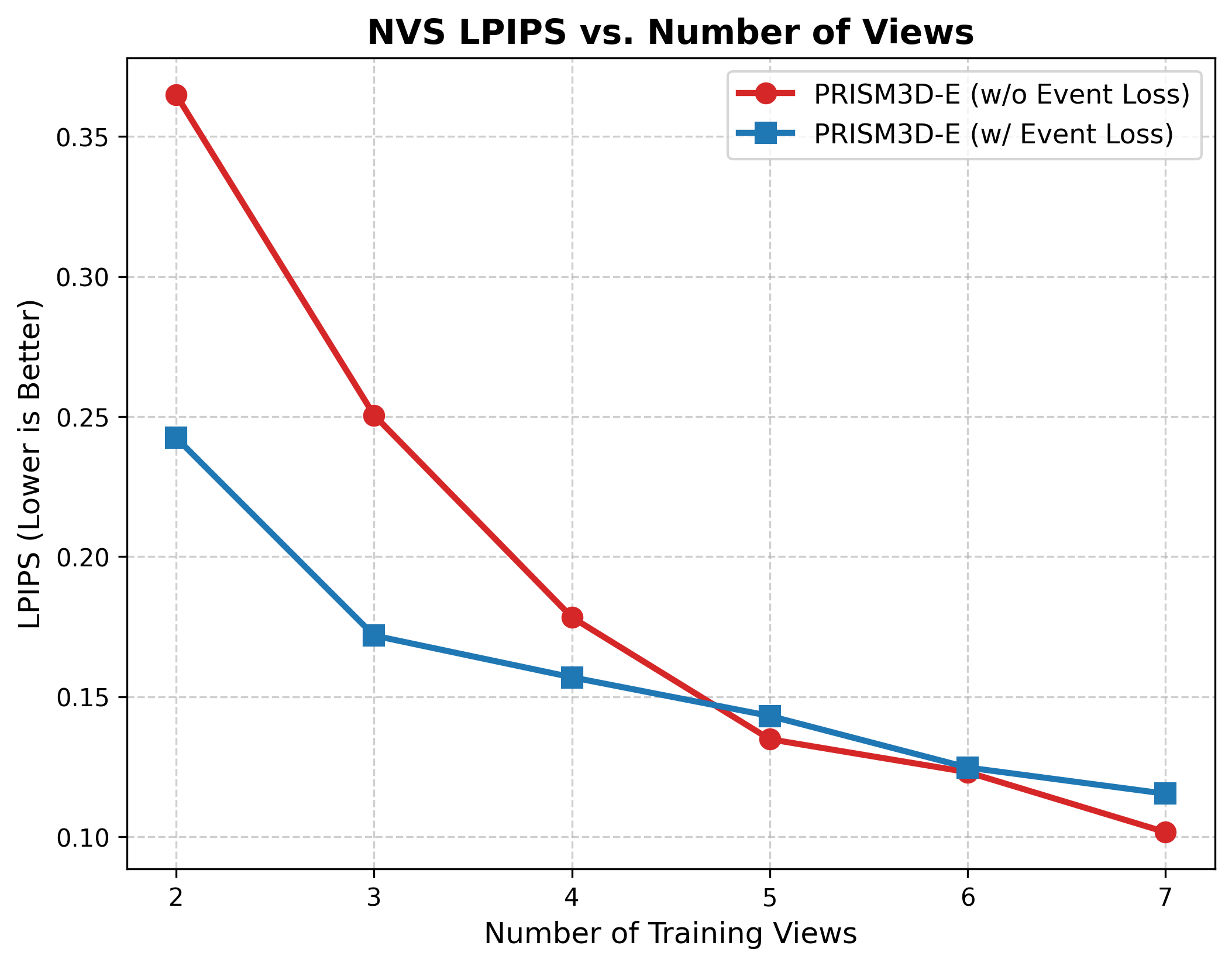}
\caption{Absolute $N$-Views}
\label{fig:n_view_sparsity}
\end{subfigure}%
\hspace{0.05\linewidth}% <-- This controls the gap between the two centered figures
\begin{subfigure}[b]{0.4\linewidth}
\centering
\includegraphics[width=\textwidth]{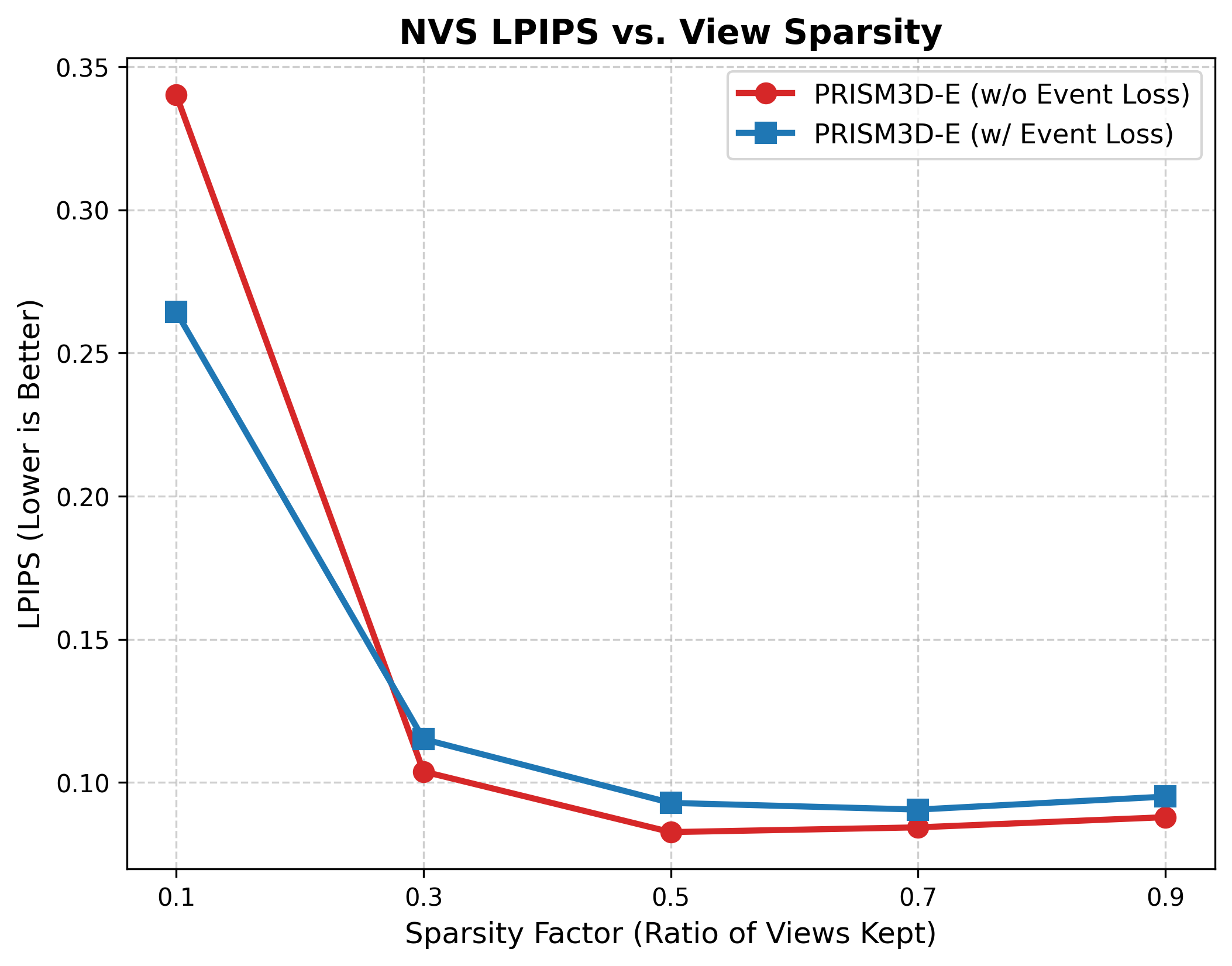}
\caption{Relative \%}
\label{fig:percent_sparsity}
\end{subfigure}
\vspace{-3mm}
\caption{Novel view synthesis LPIPS ($\downarrow$) of PRISM3D-E with and without continuous event loss under varying sparsity regimes. Left: absolute training views. Right: relative percentage.}
\label{fig:view_sparsity_analysis}
\vspace{-5mm}
\end{figure}

To further untangle this behavior, we analyze the performance of both variants across progressive degrees of view sparsity. As illustrated in Fig.~\ref{fig:view_sparsity_analysis}, continuous event supervision during optimization becomes significantly beneficial under extremely sparse-view settings (e.g., $N \le 4$) where spatial and geometric constraints from the sparse RGB images are severely limited. However, as the density of the viewpoints increases, our coupled probabilistic MCMC refinement and continuous $SE(3)$ Bézier physical trajectory modeling prove sufficiently robust to recover high-fidelity scene geometry, resulting in diminishing returns for continuous multi-modal constraints. This observation justifies our lightweight front-end architectural design choice for PRISM3D-E under standard multi-view settings.
\vspace{-9mm}
\section{Discussion}
\label{sec:discussion}
\vspace{-4mm}
In this work, we presented \textbf{PRISM3D}, a unified framework that solves the inverse problem of blind 3D scene reconstruction from extreme motion blur. We identified that the primary barrier to robust deblurring is the ``Initialization-Optimization Gap'': traditional SfM fails to register blurred images, while deep trackers yield geometry too sparse for deterministic optimization.

PRISM3D bridges this gap by coupling \textbf{Robust Deep Initialization} (via VGGSfM) with \textbf{Probabilistic Scene Modeling} (via MCMC). This synergy is critical. The deep tracker recovers the global topology that feature-based methods miss, while the probabilistic formulation enables Gaussian primitives to adaptively explore and populate the sparse geometry, preventing local minima that plague deterministic baselines. By jointly optimizing these probabilistic priors with continuous \textbf{Bézier Trajectories}, we ensure the reconstruction respects the physical integration of photons during the exposure.

Furthermore, beyond our robust standalone pipeline, we demonstrated that complementary sensor data can further extend reconstruction capabilities in severely degraded conditions. Our multi-modal extension, \textbf{PRISM3D-E}, utilizes event streams purely to secure a reliable VGGSfM initialization via intermediate pseudo-sharp image synthesis. Importantly, we preserve physical accuracy by isolating these synthesized priors from the training phase; the actual scene densification is driven entirely by the raw blurred captures and our continuous trajectory formulation. Validated on our newly introduced \textbf{PRISM3D-E Benchmark}, both configurations consistently achieve state-of-the-art performance, providing a highly versatile, event-agnostic framework for tackling extreme real-world motion blur.

\textbf{Limitations and Future Work.} While robust, PRISM3D currently assumes static scenes; extending it to dynamic environments via 4D Gaussian Splatting deformation fields is a natural next step. Additionally, fine-tuning the VGGSfM pipeline directly on extremely blurred datasets offers a clear path for immediate robustness gains. Finally, although feed-forward 3D Foundation Models (e.g., VGGT) currently introduce optimization-breaking noise, adapting them for severe degradation remains a promising long-term direction to achieve instantaneous, feed-forward scene initialization.

\begin{comment}
\textbf{Future Work:} While PRISM3D achieves robust reconstruction from sparse blurred views, we aim to push this paradigm to its theoretical limit by exploring \textbf{Single-Shot Deblurring}, recovering 3D structure from a single extremely blurred image by leveraging stronger generative priors. Additionally, we plan to extend our continuous trajectory formulation to handle \textbf{Dynamic Scenes}, where object motion and camera motion are coupled, further advancing the frontier of physics-based neural rendering.
\end{comment}
\vspace{-5mm}
\section*{Acknowledgements}
\vspace{-4mm}
We would like to acknowledge partial support from the Institute of Eminence (IoE) Research Center on VR and Haptics. KM would like to acknowledge support from the Qualcomm Faculty Award 2024.

\bibliographystyle{splncs04}
\bibliography{main}

\begin{thebibliography}{10}
\providecommand{\url}[1]{\texttt{#1}}
\providecommand{\urlprefix}{URL }
\providecommand{\doi}[1]{https://doi.org/#1}

\bibitem{barron2021mip}
Barron, J.T., Mildenhall, B., Tancik, M., Hedman, P., Martin-Brualla, R., Srinivasan, P.P.: Mip-nerf: A multiscale representation for anti-aliasing neural radiance fields. In: CVPR (2021)

\bibitem{barron2022mip}
Barron, J.T., Mildenhall, B., Verbin, D., Srinivasan, P.P., Hedman, P.: Mip-nerf 360: Unbounded anti-aliased neural radiance fields. In: CVPR (2022)

\bibitem{chen2022tensorf}
Chen, A., Xu, Z., Geiger, A., Yu, J., Su, H.: Tensorf: Tensorial radiance fields. In: ECCV (2022)

\bibitem{chen2024deblur}
Chen, W., Liu, L.: Deblur-gs: 3d gaussian splatting from camera motion blurred images. Proceedings of the ACM on Computer Graphics and Interactive Techniques  \textbf{7}(1),  1--15 (2024)

\bibitem{e2gs}
Deguchi, H., Masuda, M., Nakabayashi, T., Saito, H.: E2gs: Event enhanced gaussian splatting. In: 2024 IEEE International Conference on Image Processing (ICIP). pp. 1676--1682. IEEE (2024)

\bibitem{fridovich2022plenoxels}
Fridovich-Keil, S., Yu, A., Tancik, M., Chen, Q., Recht, B., Kanazawa, A.: Plenoxels: Radiance fields without neural networks. In: CVPR (2022)

\bibitem{jang2025splat}
Jang, H., Choi, D., Kim, D., Kang, W., Kim, M.H.: Splat-based 3d scene reconstruction with extreme motion-blur. In: Proceedings of the IEEE/CVF International Conference on Computer Vision. pp. 26425--26434 (2025)

\bibitem{jiang2022alignerf}
Jiang, Y., Hedman, P., Mildenhall, B., Xu, D., Barron, J.T., Wang, Z., Xue, T.: Alignerf: High-fidelity neural radiance fields via alignment-aware training. CVPR  (2023)

\bibitem{phototourism}
Jin, Y., Mishkin, D., Mishchuk, A., Matas, J., Fua, P., Yi, K.M., Trulls, E.: Image matching across wide baselines: From paper to practice. International Journal of Computer Vision  \textbf{129}(2),  517--547 (2021)

\bibitem{kerbl3Dgaussians}
Kerbl, B., Kopanas, G., Leimk{\"u}hler, T., Drettakis, G.: {3D Gaussian Splatting for Real-Time Radiance Field Rendering}. ACM TOG  \textbf{42}(4) (July 2023), \url{https://repo-sam.inria.fr/fungraph/3d-gaussian-splatting/}

\bibitem{kerbl20233d}
Kerbl, B., Kopanas, G., Leimk{\"u}hler, T., Drettakis, G.: 3d gaussian splatting for real-time radiance field rendering. TOG  (2023)

\bibitem{MCMC}
Kheradmand, S., Rebain, D., Sharma, G., Sun, W., Tseng, Y.C., Isack, H., Kar, A., Tagliasacchi, A., Yi, K.M.: 3d gaussian splatting as markov chain monte carlo. Advances in Neural Information Processing Systems  \textbf{37},  80965--80986 (2025)

\bibitem{lee2023exblurf}
Lee, D., Oh, J., Rim, J., Cho, S., Lee, K.M.: Exblurf: Efficient radiance fields for extreme motion blurred images. In: ICCV (2023)

\bibitem{lindell2021autoint}
Lindell, D.B., Martel, J.N., Wetzstein, G.: Autoint: Automatic integration for fast neural volume rendering. In: CVPR (2021)

\bibitem{pixel_perfect_sfm}
Lindenberger, P., Sarlin, P.E., Larsson, V., Pollefeys, M.: Pixel-perfect structure-from-motion with featuremetric refinement. In: Proceedings of the IEEE/CVF international conference on computer vision. pp. 5987--5997 (2021)

\bibitem{vggt-x}
Liu, Y., Luo, C., Tang, Z., Peng, J., Zhang, Z.: Vggt-x: When vggt meets dense novel view synthesis. arXiv preprint arXiv:2509.25191  (2025)

\bibitem{maxraymarching}
Max, N.: Optical models for direct volume rendering. IEEE Transactions on Visualization and Computer Graphics pp. 99--108 (1995). \doi{10.1109/2945.468400}

\bibitem{mildenhall2020nerf}
Mildenhall, B., Srinivasan, P.P., Tancik, M., Barron, J.T., Ramamoorthi, R., Ng, R.: {{NeRF}}: {{Representing Scenes}} as {{Neural Radiance Fields}} for {{View Synthesis}} (Aug 2020). \doi{10.48550/arXiv.2003.08934}

\bibitem{muller2022instant}
M{\"u}ller, T., Evans, A., Schied, C., Keller, A.: Instant neural graphics primitives with a multiresolution hash encoding. TOG  (2022)

\bibitem{glowmap}
Pan, L., Bar{\'a}th, D., Pollefeys, M., Sch{\"o}nberger, J.L.: Global structure-from-motion revisited. In: European Conference on Computer Vision. pp. 58--77. Springer (2024)

\bibitem{pan2019bringing}
Pan, L., Scheerlinck, C., Yu, X., Hartley, R., Liu, M., Dai, Y.: Bringing a {{Blurry Frame Alive}} at {{High Frame-Rate With}} an {{Event Camera}}. In: Proceedings of the {{IEEE}}/{{CVF Conference}} on {{Computer Vision}} and {{Pattern Recognition}}. pp. 6820--6829 (2019)

\bibitem{paszke2019pytorch}
Paszke, A., Gross, S., Massa, F., Lerer, A., Bradbury, J., Chanan, G., Killeen, T., Lin, Z., Gimelshein, N., Antiga, L., et~al.: {Pytorch: An Imperative Style, High-performance Deep Learning Library}. Advances in neural information processing systems  \textbf{32} (2019), \url{https://pytorch.org/}

\bibitem{qi2023e2nerf}
Qi, Y., Zhu, L., Zhang, Y., Li, J.: {{E2NeRF}}: {{Event Enhanced Neural Radiance Fields}} from {{Blurry Images}}. In: Proceedings of the {{IEEE}}/{{CVF International Conference}} on {{Computer Vision}}. pp. 13254--13264 (2023)

\bibitem{ebadnerf}
Qi, Y., Zhu, L., Zhao, Y., Bao, N., Li, J.: Deblurring neural radiance fields with event-driven bundle adjustment. In: Proceedings of the 32nd ACM International Conference on Multimedia. pp. 9262--9270 (2024)

\bibitem{rebecq2018esim}
Rebecq, H., Gehrig, D., Scaramuzza, D.: {{ESIM}}: An {{Open Event Camera Simulator}}. In: Proceedings of {{The}} 2nd {{Conference}} on {{Robot Learning}}. pp. 969--982. {PMLR} (Oct 2018)

\bibitem{Reiser2021ICCV}
Reiser, C., Peng, S., Liao, Y., Geiger, A.: Kilonerf: Speeding up neural radiance fields with thousands of tiny mlps. In: ICCV (2021)

\bibitem{co3d}
Reizenstein, J., Shapovalov, R., Henzler, P., Sbordone, L., Labatut, P., Novotny, D.: Common objects in 3d: Large-scale learning and evaluation of real-life 3d category reconstruction. In: Proceedings of the IEEE/CVF international conference on computer vision. pp. 10901--10911 (2021)

\bibitem{hloc}
Sarlin, P.E., Cadena, C., Siegwart, R., Dymczyk, M.: From coarse to fine: Robust hierarchical localization at large scale. In: CVPR (2019)

\bibitem{superglue}
Sarlin, P.E., DeTone, D., Malisiewicz, T., Rabinovich, A.: {SuperGlue}: Learning feature matching with graph neural networks. In: CVPR (2020)

\bibitem{colmap}
Schonberger, J.L., Frahm, J.M.: {Structure-from-motion Revisited}. In: CVPR (2016), \url{https://github.com/colmap/colmap}

\bibitem{eth3d}
Schops, T., Schonberger, J.L., Galliani, S., Sattler, T., Schindler, K., Pollefeys, M., Geiger, A.: A multi-view stereo benchmark with high-resolution images and multi-camera videos. In: Proceedings of the IEEE conference on computer vision and pattern recognition. pp. 3260--3269 (2017)

\bibitem{sun2022dvgo}
Sun, C., Sun, M., Chen, H.T.: Direct voxel grid optimization: Super-fast convergence for radiance fields reconstruction. CVPR  (2022)

\bibitem{vggt}
Wang, J., Chen, M., Karaev, N., Vedaldi, A., Rupprecht, C., Novotny, D.: Vggt: Visual geometry grounded transformer. In: Proceedings of the Computer Vision and Pattern Recognition Conference. pp. 5294--5306 (2025)

\bibitem{wang2024vggsfm}
Wang, J., Karaev, N., Rupprecht, C., Novotny, D.: Vggsfm: Visual geometry grounded deep structure from motion. In: Proceedings of the IEEE/CVF conference on computer vision and pattern recognition. pp. 21686--21697 (2024)

\bibitem{wang2023badnerf}
Wang, P., Zhao, L., Ma, R., Liu, P.: {BAD-NeRF: Bundle Adjusted Deblur Neural Radiance Fields}. In: CVPR (2023), \url{https://wangpeng000.github.io/BAD-NeRF/}

\bibitem{wu2021diver}
Wu, L., Lee, J.Y., Bhattad, A., Wang, Y., Forsyth, D.: Diver: Real-time and accurate neural radiance fields with deterministic integration for volume rendering (2022)

\bibitem{gsplat}
Ye, V., Li, R., Kerr, J., Turkulainen, M., Yi, B., Pan, Z., Seiskari, O., Ye, J., Hu, J., Tancik, M., et~al.: gsplat: An open-source library for gaussian splatting. Journal of Machine Learning Research  \textbf{26}(34),  1--17 (2025)

\bibitem{yu2021plenoctrees}
Yu, A., Li, R., Tancik, M., Li, H., Ng, R., Kanazawa, A.: Plenoctrees for real-time rendering of neural radiance fields. In: ICCV (2021)

\bibitem{yu2025evagaussians}
Yu, W., Feng, C., Li, J., Tang, J., Yang, J., Tang, Z., Cao, M., Jia, X., Yang, Y., Yuan, L., et~al.: Evagaussians: Event stream assisted gaussian splatting from blurry images. In: Proceedings of the IEEE/CVF International Conference on Computer Vision. pp. 24780--24790 (2025)

\bibitem{zamir2022restormer}
Zamir, S.W., Arora, A., Khan, S., Hayat, M., Khan, F.S., Yang, M.H.: Restormer: Efficient transformer for high-resolution image restoration. In: CVPR (2022)

\bibitem{zamir2021multi}
Zamir, S.W., Arora, A., Khan, S., Hayat, M., Khan, F.S., Yang, M.H., Shao, L.: {Multi-stage Progressive Image Restoration}. In: CVPR (2021), \url{https://github.com/swz30/MPRNet}

\bibitem{kaizhang2020nerfplusplus}
Zhang, K., Riegler, G., Snavely, N., Koltun, V.: {NeRF++}: Analyzing and improving neural radiance fields. arXiv:2010.07492  (2020)

\bibitem{zhao2024badgaussians}
Zhao, L., Wang, P., Liu, P.: {BAD-Gaussians: Bundle Adjusted Deblur Gaussian Splatting}. In: ECCV. Springer (2024)

\end{thebibliography}

% ====================================================================
% SUPPLEMENTARY MATERIAL TRANSITION
% ====================================================================
\clearpage
\appendix

% 1. Reset ONLY the section counter
% (Pages, figures, tables, and equations continue running uninterrupted!)
\setcounter{section}{0}

% 2. Switch sections to A, B, C...
\renewcommand{\thesection}{\Alph{section}}

% 3. Supplementary Title Banner
\begin{center}
    \Large \textbf{PRISM3D: Probabilistic Refinement and Robust Initialization for Physically Consistent Scene Modeling under Extreme Motion Blur}\\[10pt]
    \large \textbf{--- Supplementary Material ---}
\end{center}
\vspace{15pt}

\section{Preliminaries}
\label{sec:preliminaries}

Our proposed framework, PRISM3D, uniquely bridges robust deep tracking with probabilistic continuous scene rendering. To provide a complete theoretical foundation for our methodology, we first briefly review the core formulations of the two established frameworks that bootstrap our pipeline: VGGSfM~\cite{wang2024vggsfm} for deep tracking initialization, and 3DGS-MCMC~\cite{MCMC} for probabilistic scene refinement.

\subsection{Deep Point Tracking and Initialization (VGGSfM)}
Given a set of severely motion-blurred images, traditional feature-matching and classical incremental Structure-from-Motion (SfM) pipelines reliably fail to register cameras. To bypass this, we leverage the VGGSfM framework to establish our initial global topology. 

Unlike traditional methods that rely on chained pairwise matching, VGGSfM employs a feed-forward, coarse-to-fine Transformer architecture. It extracts reliable 2D point tracks across all frames simultaneously by constructing and flattening a multi-resolution cost-volume pyramid. Crucially for our extreme degradation setting, the tracker estimates aleatoric uncertainty (variance) for each 2D track point. This allows the system to confidently filter out outlier correspondences caused by severe motion blur. 

These robust 2D tracks, along with a deep camera pose prior, are fed into a fully differentiable Levenberg-Marquardt Bundle Adjustment (BA) layer. The BA minimizes the reprojection error to output a set of initial camera poses $\mathbf{T}$ and a 3D point cloud $\mathbf{P}$. While this global initialization successfully bypasses the failure points of classical SfM, the resulting point cloud $\mathbf{P}$ is inherently sparse, serving as the structural skeleton that our subsequent probabilistic module must densify.

\subsection{Probabilistic Refinement (3DGS-MCMC)}
To model the physical integration of light during an exposure, PRISM3D renders latent sharp images along a continuous trajectory. We achieve this rendering via standard 3D Gaussian Splatting (3DGS)~\cite{kerbl20233d}. For a virtual camera pose $P_i$, the discrete color $C$ of a pixel coordinate $\mathbf{u}$ is computed via depth-sorted $\alpha$-blending:
\begin{equation}
    C(\mathbf{u}) = \sum_{j=1}^N c_j \alpha_j(\mathbf{u}) \left[ \prod_{m=1}^{j-1} (1 - \alpha_m(\mathbf{u})) \right],
    \label{eq:gscompose}
\end{equation}
where $c_j$ is the view-dependent color of the $j$-th Gaussian. The pixel-space opacity $\alpha_j(\mathbf{u})$ is obtained by projecting the 3D Gaussian onto the 2D image plane:
\begin{align}
    \alpha_j(\mathbf{u}) = 
    o_j \exp \Bigl(-\frac{1}{2}(\mathbf{u} - \pi(\mu_j; P_i))^\mathrm{T}\Sigma_{2D}^{-1}(\mathbf{u} - \pi(\mu_j; P_i))\Bigr),
    \label{eq:alpha}
\end{align}
where $o_j$ is the 3D opacity, $\mu_j$ is the 3D center, $\pi$ is the camera projection, and $\Sigma_{2D}$ is the projected 2D covariance.

\textbf{Stochastic Gradient Langevin Dynamics (SGLD):} 
Standard 3DGS relies on deterministic, gradient-based heuristics (cloning and splitting) to densify the scene. However, when initialized with the highly sparse point cloud $\mathbf{P}$ generated by VGGSfM, these heuristics fail to traverse the large empty spatial gaps, leading to severe local minima. 

Therefore, we adopt the 3DGS-MCMC formulation, which treats the Gaussian parameters $\Theta$ as samples drawn from an underlying scene distribution. During the backward pass, while parameters like opacity and color receive standard gradients, the spatial locations ($\mu$) are updated using SGLD. This injects controlled exploration noise $\epsilon$ into the spatial gradients derived from our continuous photometric blur loss:
\begin{equation}
\mu_{t+1} \leftarrow \mu_t 
- 
\lambda_{\text{lr}} \nabla_{\mu} \mathcal{L}(\mu_t)
+
\epsilon
\label{eq:sgldupdt}
\end{equation}
To prevent catastrophic geometric collapse and preserve the VGGSfM skeletal structure, the noise $\epsilon$ is strictly conditioned on the covariance $\Sigma$ and opacity $o$ of each individual Gaussian:
\begin{equation}
    \epsilon = \lambda_{\text{lr}} \cdot \sigma \bigl(-k(\tau - o)\bigr) \cdot \Sigma \eta \quad \text{where} \quad \eta \sim \mathcal{N}(\mathbf{0}, \mathbf{I})
    \label{eq:noise}
\end{equation}
Here, $\lambda_{\text{lr}}$ is the learning rate, $\sigma$ is the sigmoid function, and $\tau=0.005$ is a threshold parameter with steepness $k=100$. This specific formulation ensures that Gaussians explore empty space strictly along their principal axes of variance ($\Sigma$), while the sigmoid function dampens the noise for opaque, well-converged primitives. Consequently, this probabilistic mechanism robustly densifies the sparse VGGSfM priors without relying on brittle deterministic cloning.
\newpage
\vspace{-6mm}
\section{The PRISM3D-E Benchmark Dataset}
\label{sec:dataset_appendix}

While our standalone RGB \textbf{PRISM3D} pipeline is highly capable of recovering 3D scenes from extreme motion blur, we introduce the \textbf{PRISM3D-E Benchmark} to push the boundaries of reconstruction quality even further when complementary event data is available. Crucially, existing event-assisted 3D datasets primarily focus on mild degradation or standard novel view synthesis. To address the complete lack of multi-modal benchmarks dedicated to severely degenerate conditions, we build upon the extremely blurred RGB sequences originally introduced by ExBluRF~\cite{lee2023exblurf}. 

Our specific contribution is the generation of perfectly synchronized, high-speed event streams corresponding to these severe motion blur scenarios. By pairing the existing degraded RGB frames with our novel event data, we provide a comprehensive multi-modal benchmark. As detailed in the main text, leveraging these events allows us to synthesize intermediate pseudo-sharp priors to successfully bootstrap the tracking initialization, ultimately enhancing the final probabilistic optimization.

\begin{figure*}[ht]
    \centering
    \includegraphics[width=0.9\linewidth]{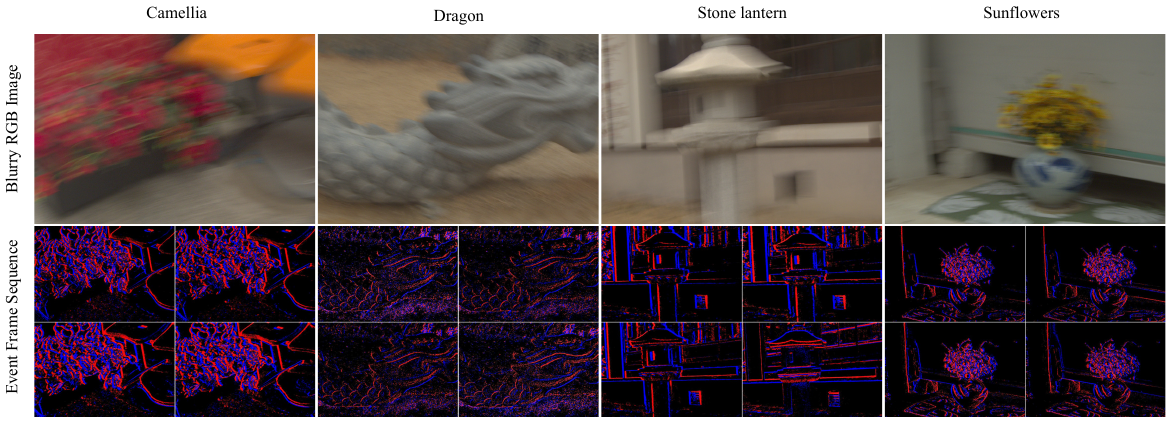}
    \caption{\textbf{The PRISM3D-E Benchmark.} We augment the challenging, motion-blurred RGB sequences from ExBluRF~\cite{lee2023exblurf} with synchronized synthetic event streams. While the high-magnitude blur in the RGB frames presents a complex optimization landscape, the high-temporal-resolution event data provides critical structural and motion priors. This multi-modal integration significantly enhances the robustness of the reconstruction, leading to superior geometric and photometric fidelity in extreme blur scenarios.}
    \label{fig:Our_Event_Dataset_Supplementary}
\end{figure*}

\newpage

\section{Algorithmic Summary of the Pipeline}
\label{sec:algorithm_details}

\begin{algorithm}[!htbp]
\small
\caption{\textsc{PRISM3D}: Pipeline for Extreme Motion Deblurring \& Inference}
\label{alg:prism3d}
% Override default Require/Ensure to standard Input/Output
\renewcommand{\algorithmicrequire}{\textbf{Input:}}
\renewcommand{\algorithmicensure}{\textbf{Output:}}

\vspace{1mm}
\noindent \hrulefill \vspace{1mm}

\noindent \textsf{\textbf{PART I: TRAINING PHASE}}
\begin{algorithmic}[1]
\Require Blurry RGB images $\{\mathbf{B}_k\}_{k=1}^{K}$, optional continuous event stream $\mathcal{E}$
\Ensure Optimized 3D Gaussians $\Theta$, continuous camera trajectories $\mathcal{T}$

\vspace{1.5mm}
\Statex \textsf{\textbf{STAGE 1: Robust Initialization}} \vspace{0.5mm}
\If{event stream $\mathcal{E}$ is available}
    \State Extract events corresponding to $\mathbf{B}_k$ to generate pseudo-sharp priors via EDI 
\EndIf
\State Run \textsf{VGGSfM} on available images to obtain initial poses $\{T_k\}$ and sparse points
\State Initialize 3D Gaussian parameters $\Theta$ from the sparse point cloud

\vspace{2mm}
\Statex \textsf{\textbf{STAGE 2: Joint Probabilistic Optimization Loop}} \vspace{0.5mm}
\While{not converged}
    \State \textbf{Sample} a blurry view $k \in \{1, \dots, K\}$ randomly
    
    \vspace{1.5mm}
    \Statex \hspace{0.4cm} \textcolor{gray}{$\triangleright$ \textit{Continuous Blur Modeling}}
    \State \textbf{Parameterize} $SE(3)$ B\'ezier trajectory $\mathcal{T}_k(t)$ for the exposure window
    \State \textbf{Sample} $n$ latent sharp poses $\{P_i\}_{i=1}^n$ uniformly along $\mathcal{T}_k(t)$
    
    \vspace{1.5mm}
    \Statex \hspace{0.4cm} \textcolor{gray}{$\triangleright$ \textit{Physical Blur Formation}}
    \State \textbf{Render} sharp latent images $\{\mathbf{C}_i\}_{i=1}^{n}$ from $\Theta$ at poses $\{P_i\}$
    \State \textbf{Approximate} the blurry image: $\mathbf{B}_{render} \approx \frac{1}{n}\sum_{i=1}^{n} \mathbf{C}_i$
    
    \vspace{1.5mm}
    \Statex \hspace{0.4cm} \textcolor{gray}{$\triangleright$ \textit{Probabilistic MCMC Update}}
    \State \textbf{Compute} photometric loss $\mathcal{L}$ between $\mathbf{B}_{render}$ and $\mathbf{B}_k$
    \State \textbf{Update} $\Theta, \mathcal{T}$ via SGLD: $\Theta \leftarrow \Theta - \eta \nabla_\Theta \mathcal{L} + \epsilon$
\EndWhile

\vspace{1.5mm}
\State \Return Optimized 3D Gaussians $\Theta$ and continuous trajectories $\mathcal{T}$
\end{algorithmic}

\vspace{1mm}
\noindent \hrulefill \vspace{1mm}

\noindent \textsf{\textbf{PART II: INFERENCE PHASE}}
\begin{algorithmic}[1]
\Require Optimized 3D scene $\Theta$, continuous trajectories $\mathcal{T}$, target novel pose $P_{novel}$
\Ensure Sharp deblurred images, sharp novel views

\vspace{1.5mm}
\Statex \hspace{0.4cm} \textcolor{gray}{$\triangleright$ \textit{Deblurring}}
\State \textbf{Evaluate} the optimized continuous trajectory $\mathcal{T}_k(t)$ at the exposure midpoint ($t=0.5$)
\State \textbf{Render} the sharp deblurred output from $\Theta$ at pose $\mathcal{T}_k(0.5)$

\vspace{1.5mm}
\Statex \hspace{0.4cm} \textcolor{gray}{$\triangleright$ \textit{Novel View Synthesis}}
\State \textbf{Render} a sharp novel view from the optimized 3D scene $\Theta$ at target camera pose $P_{novel}$

\vspace{1.5mm}
\State \Return Sharp deblurred images, sharp novel views
\end{algorithmic}
\vspace{1mm}
\end{algorithm}

\clearpage

\section{Experiments}
\subsection{Quantitative Results}

In this section, we present quantitative evaluations on the PRISM3D-E Synthetic Benchmark for two main tasks: sharp novel view synthesis (Table \ref{tab:quan_table}) and image deblurring (Table \ref{tab:quant_deblur_table_appendix}). To ensure a fair comparison, we categorize the evaluated methods into three groups based on their input modality: (1) \textbf{Blind RGB} (w/o Events), (2) \textbf{Event-Assisted} (w/ Events), and (3) \textbf{Oracle Baselines} (w/ Sharp Supervision). Note that the oracle methods rely on impractical sharp-image initialization and are included strictly for reference.

\begin{table}[!ht]
    \centering
    \caption{\textbf{Quantitative comparisons for sharp novel view synthesis (deblurring + novel view synthesis) on the PRISM3D-E Benchmark (Synthetic).} 
The table is grouped by input modality: (1) \textbf{Blind RGB} (w/o Events), (2) \textbf{Event-Assisted} (w/ Events), and (3) \textbf{Oracle Baselines} (w/ Sharp Supervision). Group 3 methods (ExBluRF*, BAD-Gaussians*, Deblur-GS*) are included only for reference as they rely on impractical sharp-image initialization. Best and second-best results within ranked groups are highlighted in \colorbox{green!25}{green} and \colorbox{red!25}{orange}.}
    \label{tab:quan_table}
    \renewcommand{\arraystretch}{1.35}
    \resizebox{0.95\linewidth}{!}{
        \begin{tabular}{c||c||ccc|cccc|ccc}
            \toprule
            & & \multicolumn{3}{c|}{\textbf{Blind RGB (w/o Events)}} & \multicolumn{4}{c|}{\textbf{Event-Assisted (w/ Events)}} & \multicolumn{3}{c}{\textbf{Oracle Baselines(*)}} \\
            \textbf{Scene} & \textbf{Metric} & \textbf{MPRNet} & \textbf{Restormer} & \textbf{PRISM3D} & \textbf{EDI+3DGS} & \textbf{E2NeRF} & \textbf{EBAD-NeRF} & \textbf{PRISM3D-E} & \textbf{ExBluRF*} & \textbf{BAD-Gaussians*} & \textbf{Deblur-GS*} \\
            \hhline{=::=::===::====::===}
            \multirow{3}{*}{\textbf{Bench}} 
                & PSNR$\uparrow$ & 25.35 & \colorbox{red!25}{26.39} & \colorbox{green!25}{29.86} & \colorbox{red!25}{28.95} & {25.41} & {28.15} & \colorbox{green!25}{33.55} & 31.93 & 32.54 & 33.26 \\
                & SSIM$\uparrow$ & 0.678 & \colorbox{red!25}{0.720} & \colorbox{green!25}{0.841} &  \colorbox{red!25}{0.865} & {0.708} &{0.822} & \colorbox{green!25}{0.924} & 0.877 & 0.901 & 0.929 \\
                & LPIPS$\downarrow$ & 0.425 & \colorbox{red!25}{0.356} & \colorbox{green!25}{0.118} & 0.201 & {0.438} & \colorbox{red!25}{0.172} & \colorbox{green!25}{0.063} & 0.111 & 0.046 & 0.116 \\
            \hhline{-||-||----------}
            \multirow{3}{*}{\textbf{Camellia}} 
                & PSNR$\uparrow$ & 24.84 & \colorbox{red!25}{25.14} & \colorbox{green!25}{28.56} & 22.46 & \colorbox{red!25}{28.07} & 24.33 & \colorbox{green!25}{29.47} & 28.02 & 28.83 & 29.14 \\
                & SSIM$\uparrow$ & 0.669 & \colorbox{red!25}{0.690} & \colorbox{green!25}{0.821} & \colorbox{red!25}{0.762} & {0.721} & {0.743} & \colorbox{green!25}{0.873} & 0.715 & 0.815 & 0.874 \\
                & LPIPS$\downarrow$ & 0.395 & \colorbox{red!25}{0.351} & \colorbox{green!25}{0.129} & 0.271 & {0.329} & \colorbox{red!25}{0.192} & \colorbox{green!25}{0.108} & 0.313 & 0.099 & 0.158 \\
            \hhline{-||-||----------}
            \multirow{3}{*}{\textbf{Dragon}} 
                & PSNR$\uparrow$ & \colorbox{red!25}{29.96} & {28.37} & \colorbox{green!25}{32.43} & 33.27 & {30.89} & \colorbox{red!25}{33.99} & \colorbox{green!25}{37.01} & 33.45 & 36.98 & 35.25 \\
                & SSIM$\uparrow$ & \colorbox{red!25}{0.731} & {0.704} & \colorbox{green!25}{0.818} & 0.842 & {0.697} & \colorbox{red!25}{0.864} & \colorbox{green!25}{0.925} & 0.828 & 0.930 & 0.879 \\
                & LPIPS$\downarrow$ & \colorbox{red!25}{0.454} & {0.465} & \colorbox{green!25}{0.171} & 0.243 & {0.433} & \colorbox{red!25}{0.202} & \colorbox{green!25}{0.069} & 0.180 & 0.045 & 0.245 \\
            \hhline{-||-||----------}
            \multirow{3}{*}{\textbf{Jars}} 
                & PSNR$\uparrow$ & 25.36 & \colorbox{red!25}{25.57} & \colorbox{green!25}{31.42} & 28.13 & \colorbox{red!25}{29.85} & 28.89 & \colorbox{green!25}{32.35} & 30.85 & 31.52 & 31.74 \\
                & SSIM$\uparrow$ & 0.680 & \colorbox{red!25}{0.687} & \colorbox{green!25}{0.879} & 0.831 & {0.775} & \colorbox{red!25}{0.838} & \colorbox{green!25}{0.898} & 0.840 & 0.867 & 0.895 \\
                & LPIPS$\downarrow$ & 0.406 & \colorbox{red!25}{0.371} & \colorbox{green!25}{0.127} & 0.238 & {0.334} & \colorbox{red!25}{0.198} & \colorbox{green!25}{0.108} & 0.156 & 0.078 & 0.172 \\
            \hhline{-||-||----------}
            \multirow{3}{*}{\textbf{Jars2}} 
                & PSNR$\uparrow$ & 24.33 & \colorbox{red!25}{26.43} & \colorbox{green!25}{28.14} & 24.74 & \colorbox{red!25}{27.71} & 27.39 & \colorbox{green!25}{28.79} & 30.89 & 28.94 & 28.90 \\
                & SSIM$\uparrow$ & 0.745 & \colorbox{red!25}{0.814} & \colorbox{green!25}{0.873} & 0.812 & {0.770} & \colorbox{red!25}{0.863} & \colorbox{green!25}{0.906} & 0.860 & 0.851 & 0.890 \\
                & LPIPS$\downarrow$ & 0.358 & \colorbox{red!25}{0.275} & \colorbox{green!25}{0.173} & 0.262 & {0.383} & \colorbox{red!25}{0.171} & \colorbox{green!25}{0.133} & 0.113 & 0.114 & 0.218 \\
            \hhline{-||-||----------}
            \multirow{3}{*}{\textbf{Postbox}} 
                & PSNR$\uparrow$ & 25.89 & \colorbox{red!25}{26.52} & \colorbox{green!25}{27.74} & 24.99 & \colorbox{red!25}{30.66} & 26.82 & \colorbox{green!25}{31.33} & 31.40 & 26.40 & 30.35 \\
                & SSIM$\uparrow$ & 0.736 & \colorbox{red!25}{0.753} & \colorbox{green!25}{0.788} & 0.789 & {0.813} & \colorbox{red!25}{0.826} & \colorbox{green!25}{0.906} & 0.864 & 0.757 & 0.888 \\
                & LPIPS$\downarrow$ & 0.318 & \colorbox{red!25}{0.286} & \colorbox{green!25}{0.150} & 0.228 & {0.262} & \colorbox{red!25}{0.151} & \colorbox{green!25}{0.070} & 0.095 & 0.123 & 0.162 \\
            \hhline{-||-||----------}
            \multirow{3}{*}{\textbf{Stone Lantern}} 
                & PSNR$\uparrow$ & 24.97 & \colorbox{red!25}{26.68} & \colorbox{green!25}{28.29} & 26.48 & \colorbox{green!25}{30.47} & 26.29 & \colorbox{red!25}{29.43} & 28.24 & 28.29 & 28.59 \\
                & SSIM$\uparrow$ & 0.785 & \colorbox{red!25}{0.831} & \colorbox{green!25}{0.849} & 0.825 & \colorbox{red!25}{0.836} & 0.802 & \colorbox{green!25}{0.894} & 0.765 & 0.843 & 0.863 \\
                & LPIPS$\downarrow$ & 0.342 & \colorbox{red!25}{0.280} & \colorbox{green!25}{0.195} & 0.270 & {0.324} & \colorbox{red!25}{0.264} & \colorbox{green!25}{0.152} & 0.236 & 0.143 & 0.236 \\
            \hhline{-||-||----------}
            \multirow{3}{*}{\textbf{Sunflowers}} 
                & PSNR$\uparrow$ & 28.86 & \colorbox{green!25}{29.55} & \colorbox{red!25}{29.47} & 31.38 & \colorbox{red!25}{31.74} & 30.98 & \colorbox{green!25}{33.69} & 34.46 & 34.06 & 34.19 \\
                & SSIM$\uparrow$ & 0.837 & \colorbox{red!25}{0.847} & \colorbox{green!25}{0.854} & \colorbox{red!25}{0.914} & {0.850} & {0.903} & \colorbox{green!25}{0.938} & 0.920 & 0.942 & 0.943 \\
                & LPIPS$\downarrow$ & 0.242 & \colorbox{red!25}{0.206} & \colorbox{green!25}{0.163} & 0.144 & {0.310} & \colorbox{red!25}{0.117} & \colorbox{green!25}{0.077} & 0.093 & 0.065 & 0.116 \\
            \hhline{=::=::===::====::===}
            \multirow{3}{*}{\textbf{Average}} & PSNR$\uparrow$ & 26.19 & \colorbox{red!25}{26.83} & \colorbox{green!25}{29.49} & 27.55 & \colorbox{red!25}{29.35} & 28.36 & \colorbox{green!25}{31.95} & 31.15 & 30.95 & 31.43 \\
                & SSIM$\uparrow$ & 0.733 & \colorbox{red!25}{0.756} & \colorbox{green!25}{0.840} & 0.830 & {0.771} & \colorbox{red!25}{0.833} & \colorbox{green!25}{0.908} & 0.834 & 0.863 & 0.895 \\
                & LPIPS$\downarrow$ & 0.368 & \colorbox{red!25}{0.324} & \colorbox{green!25}{0.153} & 0.232 & {0.352} & \colorbox{red!25}{0.183} & \colorbox{green!25}{0.097} & 0.162 & 0.089 & 0.178 \\
            \bottomrule
        \end{tabular}
    }
\end{table}

\clearpage

\begin{table*}[!p]
    \centering
    \caption{\textbf{Quantitative comparisons for deblurring on the Synthetic Dataset.} 
    The table is organized into three groups: (1) \textbf{Blind RGB} (w/o Events), 
    (2) \textbf{Event-Assisted} (w/ Events), and (3) \textbf{Oracle Baselines} (w/ Sharp Supervision). Group 3 methods (ExBluRF*, BAD-Gaussians*, Deblur-GS*) are included only for reference as they rely on impractical sharp-image initialization.}
    \label{tab:quant_deblur_table_appendix}
    \renewcommand{\arraystretch}{1.35}
    \resizebox{0.95\linewidth}{!}{
        \begin{tabular}{c||c||ccc|cccc|ccc}
            \toprule
            & & \multicolumn{3}{c|}{\textbf{Blind RGB (w/o Events)}} & \multicolumn{4}{c|}{\textbf{Event-Assisted (w/ Events)}} & \multicolumn{3}{c}{\textbf{Oracle Baselines(*)}} \\
            \textbf{Scene} & \textbf{Metric} & \textbf{MPRNet} & \textbf{Restormer} & \textbf{PRISM3D (Ours)} & \textbf{EDI+3DGS} & \textbf{E2NeRF} & \textbf{EBAD-NeRF} & \textbf{PRISM3D-E (Ours)} & \textbf{ExBluRF*} & \textbf{BAD-Gaussians*} & \textbf{Deblur-GS*} \\
            \hhline{=::=::===::====::===}
            \multirow{3}{*}{\textbf{Bench}} 
                & PSNR$\uparrow$ & 25.35 & \colorbox{red!25}{26.39} & \colorbox{green!25}{30.43} & 26.23 & 26.85 & \colorbox{red!25}{28.73} & \colorbox{green!25}{34.23} & 32.58 & 33.06 & 33.57 \\
                & SSIM$\uparrow$ & 0.678 & \colorbox{red!25}{0.720} & \colorbox{green!25}{0.855} & 0.761 & 0.749 & \colorbox{red!25}{0.839} & \colorbox{green!25}{0.933} & 0.873 & 0.910 & 0.933 \\
                & LPIPS$\downarrow$ & 0.425 & \colorbox{red!25}{0.356} & \colorbox{green!25}{0.098} & 0.219 & 0.270 & \colorbox{red!25}{0.156} & \colorbox{green!25}{0.045} & 0.123 & 0.034 & 0.107 \\
            \hhline{-||-||----------}
            \multirow{3}{*}{\textbf{Camellia}} 
                & PSNR$\uparrow$ & 24.84 & \colorbox{red!25}{25.14} & \colorbox{green!25}{29.45} & 21.43 & \colorbox{red!25}{24.78} & {24.33} & \colorbox{green!25}{30.12} & 28.29 & 29.45 & 29.48 \\
                & SSIM$\uparrow$ & 0.669 & \colorbox{red!25}{0.690} & \colorbox{green!25}{0.848} & 0.667 & 0.702 & \colorbox{red!25}{0.753} & \colorbox{green!25}{0.890} & 0.744 & 0.834 & 0.885 \\
                & LPIPS$\downarrow$ & 0.395 & \colorbox{red!25}{0.351} & \colorbox{green!25}{0.110} & 0.280 & 0.232 & \colorbox{red!25}{0.175} & \colorbox{green!25}{0.092} & 0.318 & 0.086 & 0.147 \\
            \hhline{-||-||----------}
            \multirow{3}{*}{\textbf{Dragon}} 
                & PSNR$\uparrow$ & \colorbox{red!25}{29.96} & {28.37} & \colorbox{green!25}{33.09} & 32.51 & 30.36 & \colorbox{red!25}{34.37} & \colorbox{green!25}{37.48} & 33.52 & 37.69 & 35.83 \\
                & SSIM$\uparrow$ & \colorbox{red!25}{0.731} & {0.704} & \colorbox{green!25}{0.837} & 0.808 & 0.752 & \colorbox{red!25}{0.873} & \colorbox{green!25}{0.932} & 0.831 & 0.938 & 0.892 \\
                & LPIPS$\downarrow$ & \colorbox{red!25}{0.454} & {0.465} & \colorbox{green!25}{0.150} & 0.241 & 0.268 & \colorbox{red!25}{0.198} & \colorbox{green!25}{0.060} & 0.192 & 0.039 & 0.232 \\
            \hhline{-||-||----------}
            \multirow{3}{*}{\textbf{Jars}} 
                & PSNR$\uparrow$ & 25.36 & \colorbox{red!25}{25.57} & \colorbox{green!25}{32.22} & 27.12 & 26.52 & \colorbox{red!25}{29.35} & \colorbox{green!25}{33.30} & 30.50 & 32.30 & 32.09 \\
                & SSIM$\uparrow$ & 0.680 & \colorbox{red!25}{0.687} & \colorbox{green!25}{0.897} & 0.770 & 0.744 & \colorbox{red!25}{0.858} & \colorbox{green!25}{0.917} & 0.843 & 0.878 & 0.905 \\
                & LPIPS$\downarrow$ & 0.406 & \colorbox{red!25}{0.371} & \colorbox{green!25}{0.093} & 0.232 & 0.241 & \colorbox{red!25}{0.172} & \colorbox{green!25}{0.076} & 0.161 & 0.067 & 0.157 \\
            \hhline{-||-||----------}
            \multirow{3}{*}{\textbf{Jars2}} 
                & PSNR$\uparrow$ & 24.33 & \colorbox{red!25}{26.43} & \colorbox{green!25}{28.74} & 25.23 & 23.74 & \colorbox{red!25}{27.68} & \colorbox{green!25}{30.01} & 31.67 & 29.39 & 29.71 \\
                & SSIM$\uparrow$ & 0.745 & \colorbox{red!25}{0.814} & \colorbox{green!25}{0.882} & 0.786 & 0.732 & \colorbox{red!25}{0.874} & \colorbox{green!25}{0.921} & 0.901 & 0.852 & 0.902 \\
                & LPIPS$\downarrow$ & 0.358 & \colorbox{red!25}{0.275} & \colorbox{green!25}{0.129} & 0.221 & 0.313 & \colorbox{red!25}{0.151} & \colorbox{green!25}{0.090} & 0.118 & 0.091 & 0.191 \\
            \hhline{-||-||----------}
            \multirow{3}{*}{\textbf{Postbox}} 
                & PSNR$\uparrow$ & 25.89 & \colorbox{red!25}{26.52} & \colorbox{green!25}{27.48} & 24.96 & 26.90 & \colorbox{red!25}{27.24} & \colorbox{green!25}{32.22} & 31.58 & 29.87 & 30.97 \\
                & SSIM$\uparrow$ & 0.736 & \colorbox{red!25}{0.753} & \colorbox{green!25}{0.760} & 0.767 & 0.784 & \colorbox{red!25}{0.832} & \colorbox{green!25}{0.913} & 0.858 & 0.823 & 0.894 \\
                & LPIPS$\downarrow$ & 0.318 & \colorbox{red!25}{0.286} & \colorbox{green!25}{0.164} & 0.210 & 0.172 & \colorbox{red!25}{0.146} & \colorbox{green!25}{0.061} & 0.134 & 0.075 & 0.153 \\
            \hhline{-||-||----------}
            \multirow{3}{*}{\textbf{Stone Lantern}} 
                & PSNR$\uparrow$ & 24.97 & \colorbox{red!25}{26.68} & \colorbox{green!25}{28.63} & 25.55 & 25.14 & \colorbox{red!25}{27.40} & \colorbox{green!25}{30.38} & 28.24 & 29.66 & 26.74 \\
                & SSIM$\uparrow$ & 0.785 & \colorbox{red!25}{0.831} & \colorbox{green!25}{0.864} & 0.795 & 0.765 & \colorbox{red!25}{0.838} & \colorbox{green!25}{0.916} & 0.813 & 0.891 & 0.802 \\
                & LPIPS$\downarrow$ & 0.342 & \colorbox{red!25}{0.280} & \colorbox{green!25}{0.148} & 0.231 & 0.255 & \colorbox{red!25}{0.228} & \colorbox{green!25}{0.107} & 0.243 & 0.099 & 0.266 \\
            \hhline{-||-||----------}
            \multirow{3}{*}{\textbf{Sunflowers}} 
                & PSNR$\uparrow$ & 28.86 & \colorbox{red!25}{29.55} & \colorbox{green!25}{30.60} & 29.14 & 28.67 & \colorbox{red!25}{31.29} & \colorbox{green!25}{34.63} & 32.92 & 34.57 & 34.76 \\
                & SSIM$\uparrow$ & 0.837 & \colorbox{red!25}{0.847} & \colorbox{green!25}{0.873} & 0.862 & 0.840 & \colorbox{red!25}{0.907} & \colorbox{green!25}{0.947} & 0.900 & 0.947 & 0.947 \\
                & LPIPS$\downarrow$ & 0.242 & \colorbox{red!25}{0.206} & \colorbox{green!25}{0.146} & 0.156 & 0.179 & \colorbox{red!25}{0.112} & \colorbox{green!25}{0.062} & 0.122 & 0.061 & 0.116 \\
            \hhline{=::=::===::====::===}
            \multirow{3}{*}{\textbf{Average}} 
                & PSNR$\uparrow$ & 26.19 & \colorbox{red!25}{26.83} & \colorbox{green!25}{30.08} & 26.52 & 26.62 & \colorbox{red!25}{28.80} & \colorbox{green!25}{32.80} & 31.16 & 32.00 & 31.64 \\
                & SSIM$\uparrow$ & 0.733 & \colorbox{red!25}{0.756} & \colorbox{green!25}{0.852} & 0.777 & 0.758 & \colorbox{red!25}{0.847} & \colorbox{green!25}{0.921} & 0.845 & 0.884 & 0.895 \\
                & LPIPS$\downarrow$ & 0.368 & \colorbox{red!25}{0.324} & \colorbox{green!25}{0.130} & 0.224 & 0.241 & \colorbox{red!25}{0.167} & \colorbox{green!25}{0.074} & 0.176 & 0.069 & 0.171 \\
            \bottomrule
        \end{tabular}
    }
\end{table*}

\clearpage

\subsection{Qualitative Results}

\begin{figure*}[!htbp]
    \centering
    \includegraphics[width=0.99\linewidth]{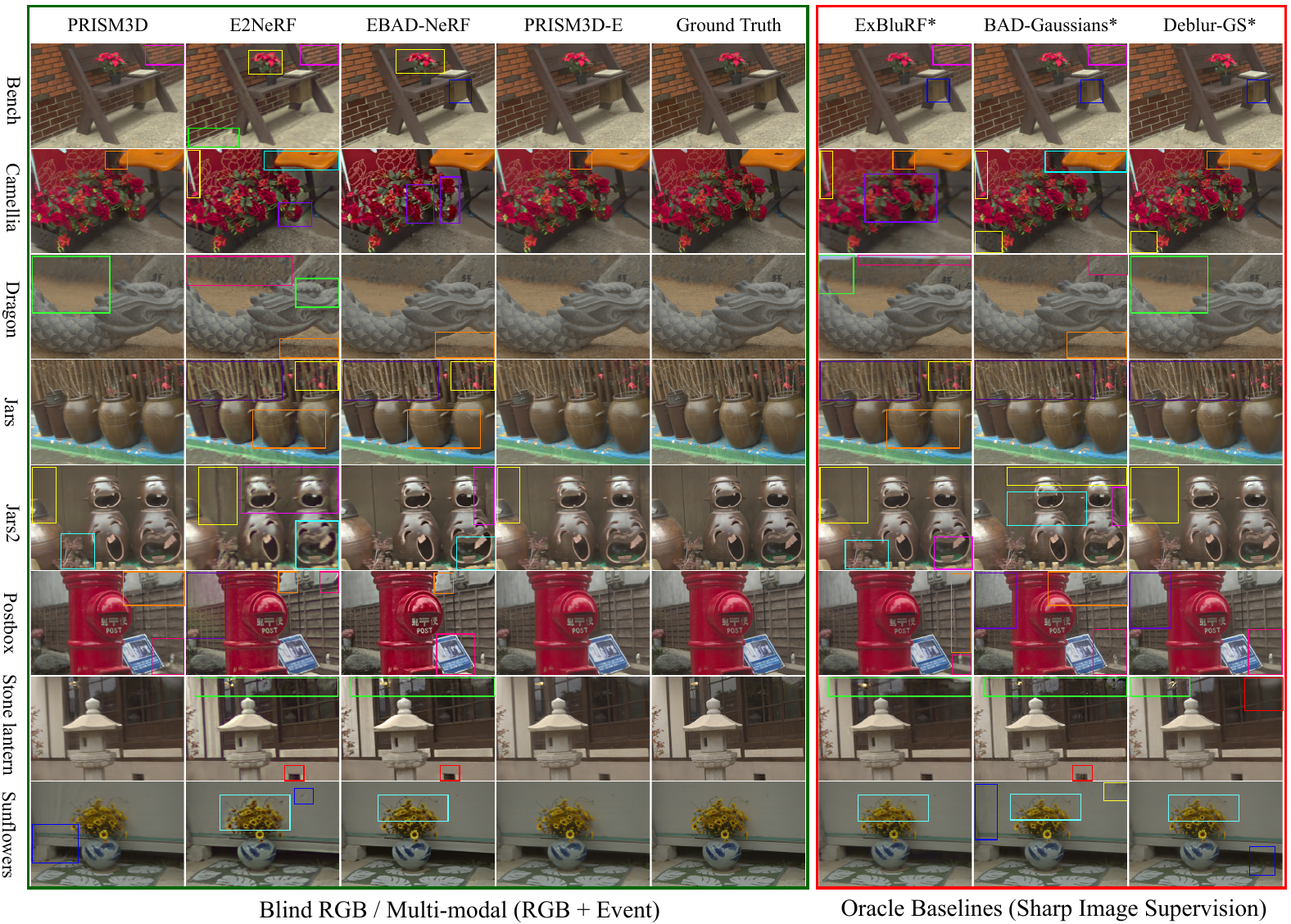}
 \caption{\textbf{Novel View Synthesis Results on the Synthetic Dataset:} Beyond restoring input views, our framework synthesizes sharp images from unseen camera angles. While existing approaches often suffer from structural drift, \textbf{PRISM3D} maintains consistency without requiring oracle initialization. Notably, our event-assisted variant \textbf{PRISM3D-E} delivers sharper details and fewer artifacts than even the oracle baselines, effectively resolving color bleeding under extreme blur.} 
    \label{fig:comparisons_synth_nvs_appendix}
\end{figure*}

We present qualitative comparisons of our proposed PRISM3D framework against state-of-the-art methods. Figures \ref{fig:comparisons_synth} and \ref{fig:comparisons_real} demonstrate the image deblurring performance on the synthetic and real-world datasets, respectively. Across both domains, our method consistently restores fine structural details and textures from severely motion-blurred inputs while minimizing visual artifacts. Furthermore, Figure \ref{fig:comparisons_synth_nvs_appendix} visualizes the results for sharp novel view synthesis on the synthetic dataset. Note that novel view synthesis evaluations are omitted for the real-world dataset due to the inherent absence of corresponding sharp ground-truth views.

\paragraph{Deblurring Comparisons}

\begin{figure}[!htb]
    \centering
    \vspace{-4mm}
    \includegraphics[width=0.99\linewidth]{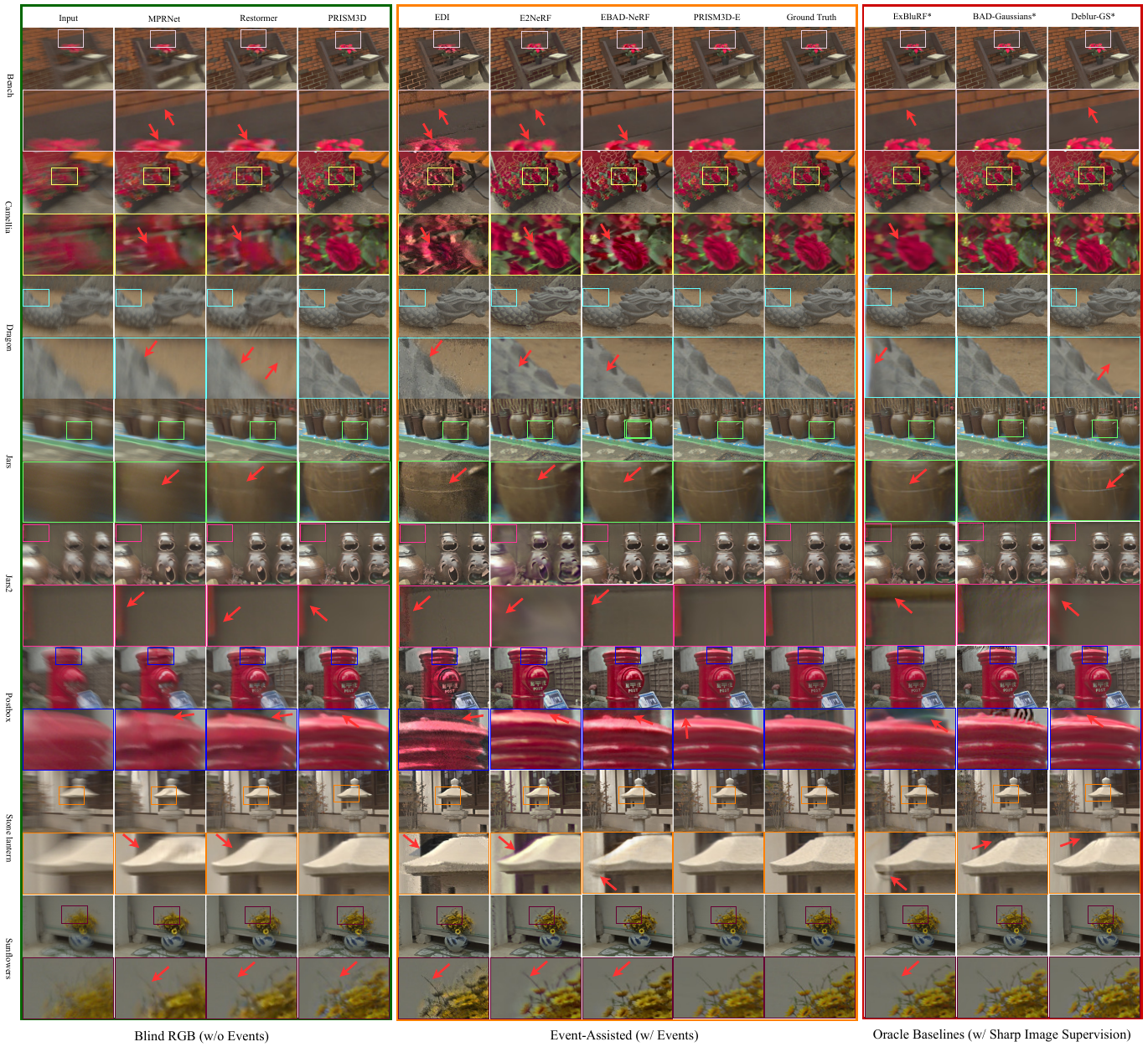}
 \caption{\textbf{Qualitative Comparisons on the Synthetic Dataset:} While oracle methods (denoted by *) rely on impractical sharp-image initialization, our blind framework \textbf{PRISM3D} achieves superior visual clarity and structural integrity, outperforming even the event-assisted baselines. Furthermore, our multi-modal variant \textbf{PRISM3D-E} surpasses the oracle baselines, effectively resolving extreme motion blur and eliminating the residual ghosting and artifacts prevalent in existing state-of-the-art methods.}
    \label{fig:comparisons_synth}
    \vspace{-4mm}
\end{figure}

\clearpage

\begin{figure}[!p]
    \centering
    %\vspace{-4mm}
    \includegraphics[width=0.95\linewidth]{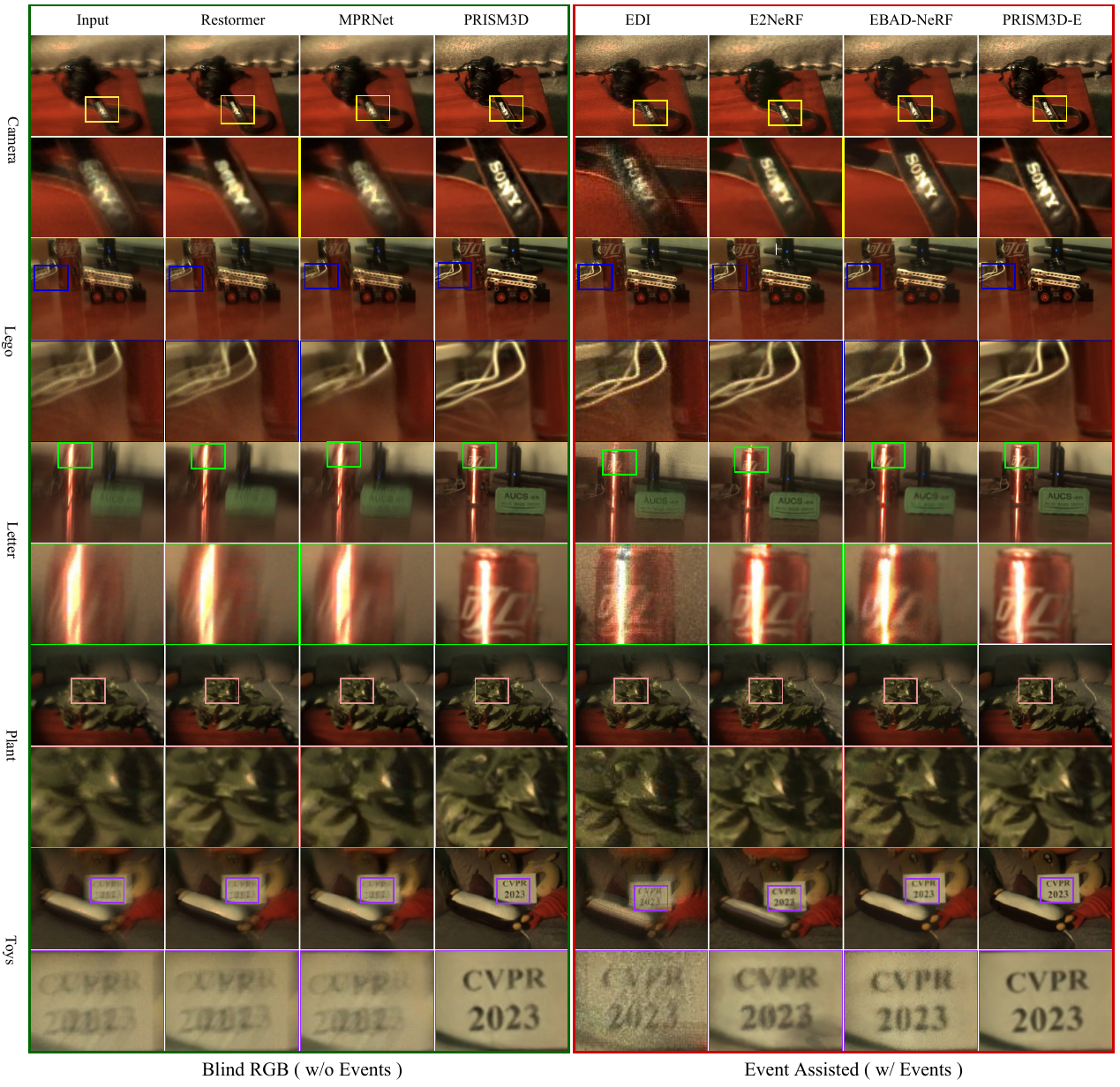}
 \caption{\textbf{Deblurring Results on the Real Dataset:} Performance on in-the-wild captures featuring complex, unknown camera trajectories. As highlighted in the zoomed-in insets, PRISM3D successfully restores illegible text and intricate textures. In contrast, existing methods tend to over-smooth the outputs or amplify real-world sensor noise.}
    \label{fig:comparisons_real}
    \vspace{-6mm}
\end{figure}

\clearpage

\paragraph{Sharp Novel View Synthesis Comparisons}

\clearpage

\subsection{Ablations: Component-wise Analysis for Deblurring} 
We ablate the components of PRISM3D and PRISM3D-E for deblurring on the synthetic dataset (Table~\ref{tab:deblur_synthetic_avg}). Similar to our novel view synthesis findings, HLOC lags behind VGGSfM by 0.90 dB in PSNR. Furthermore, 3D foundation models like VGGT-X struggle under extreme blur, underperforming our initialization by 1.37 dB. This reinforces VGGSfM's robust capability to extract global topology from severely degraded inputs. 

\begin{table}[!ht]
    \centering
    \caption{\textbf{Ablation study for deblurring on the Synthetic Dataset.} Average metrics across all 8 synthetic scenes. The inclusion of each module (VGGSfM, MCMC, EDI) yields synergistic improvements. Best and second-best results within each input modality are highlighted in \colorbox{green!25}{green} and \colorbox{red!25}{orange}.}
    \label{tab:deblur_synthetic_avg}
    \footnotesize
    \renewcommand{\arraystretch}{1.1}
    \setlength{\tabcolsep}{4pt}
    \setlength{\fboxsep}{0pt}

    \begin{tabular}{@{}l cccc ccc@{}}
        \toprule
        \multirow{2}{*}{\textbf{Variant}} & \multicolumn{4}{c}{\textbf{Configuration}} & \multicolumn{3}{c}{\textbf{Metrics}} \\
        \cmidrule(lr){2-5} \cmidrule(l){6-8}
        & \textbf{SfM} & \textbf{MCMC} & \textbf{Traj.} & \textbf{EDI} & \textbf{PSNR$\uparrow$} & \textbf{SSIM$\uparrow$} & \textbf{LPIPS$\downarrow$} \\
        \midrule

        w/ HLOC  & HLOC   & \checkmark & Bézier & $\times$ & 29.18 & 0.794 & \colorbox{red!25}{0.124} \\
        w/ VGGT-X& VGGT-X & \checkmark & Bézier & $\times$ & 28.71 & 0.773 & 0.164 \\
        w/o MCMC & VGGSfM & $\times$   & Bézier & $\times$ & \colorbox{red!25}{29.60} & 0.822 & \colorbox{green!25}{0.123} \\
        w/ Linear& VGGSfM & \checkmark & Linear & $\times$ & 28.14 & 0.800 & 0.170 \\
        w/ Spline& VGGSfM & \checkmark & Spline & $\times$ & 29.47 & \colorbox{red!25}{0.840} & 0.130 \\
        \rowcolor{gray!15}
        \textbf{PRISM3D} & VGGSfM & \checkmark & Bézier & $\times$ & \colorbox{green!25}{30.08} & \colorbox{green!25}{0.852} & 0.130 \\

        \midrule

        w/ COLMAP& COLMAP & \checkmark & Bézier & \checkmark & 32.00 & 0.899 & 0.113 \\
        w/o MCMC & VGGSfM & $\times$   & Bézier & \checkmark & 32.00 & 0.884 & \colorbox{green!25}{0.069} \\
        w/ Linear& VGGSfM & \checkmark & Linear & \checkmark & 30.42 & 0.870 & 0.120 \\
        w/ Spline& VGGSfM & \checkmark & Spline & \checkmark & \colorbox{red!25}{32.05} & \colorbox{red!25}{0.910} & 0.080 \\
        \rowcolor{gray!15}
        \textbf{PRISM3D-E}& VGGSfM & \checkmark & Bézier & \checkmark & \colorbox{green!25}{32.80} & \colorbox{green!25}{0.921} & \colorbox{red!25}{0.074} \\

        \bottomrule
    \end{tabular}
\end{table}

MCMC-based probabilistic refinement proves crucial, adding 0.48 dB in PSNR (standalone) and 0.80 dB (with events). We also validate our physics-consistent blur formation model: assuming a naive linear trajectory drops PSNR by 1.94 dB. Our continuous $SE(3)$ Bézier formulation outperforms standard spline interpolation by 0.61 dB in the blind RGB setting. With high-temporal-resolution event priors (EDI) integrated, accurate trajectory modeling is even more vital; Bézier outperforms linear and spline interpolations by 2.38 dB and 0.75 dB, respectively. Overall, integrating event priors yields the largest boost, improving PSNR by 2.72 dB over standalone PRISM3D.

\clearpage

\subsection{Qualitative Ablation Study}

\begin{figure*}[!htbp] % <--- [p] forces it to a dedicated, vertically centered page
    \centering
    % --- First Figure ---
    \includegraphics[width=0.99\linewidth]{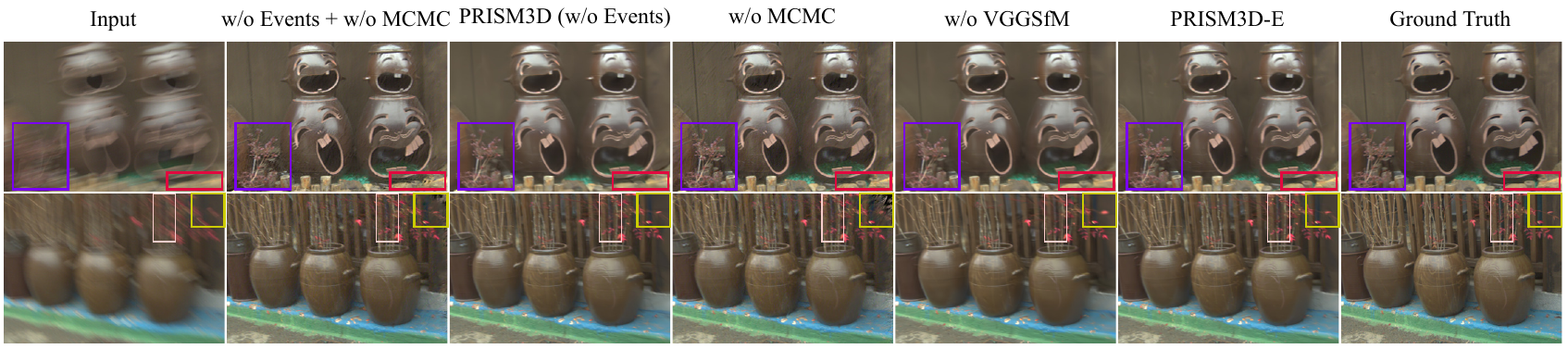}
    \caption{\textbf{Qualitative ablation study on the PRISM3D-E Benchmark (Synthetic).} We visually assess the contribution of different modules within our framework. Our base RGB method (\textbf{PRISM3D}) achieves strong deblurring, with the MCMC module playing a crucial role in reducing floaters and background artifacts. When complementary event data is utilized to bootstrap initialization (\textbf{PRISM3D-E}), the reconstructions become even sharper, recovering fine text and high-frequency structural details that are otherwise lost in the severely degraded inputs.}
    \label{fig:ablations_synthetic}
    
    \vspace{8mm} % <--- Adjust this value to control the gap between the two figures
    
    % --- Second Figure ---
    \includegraphics[width=0.9\linewidth]{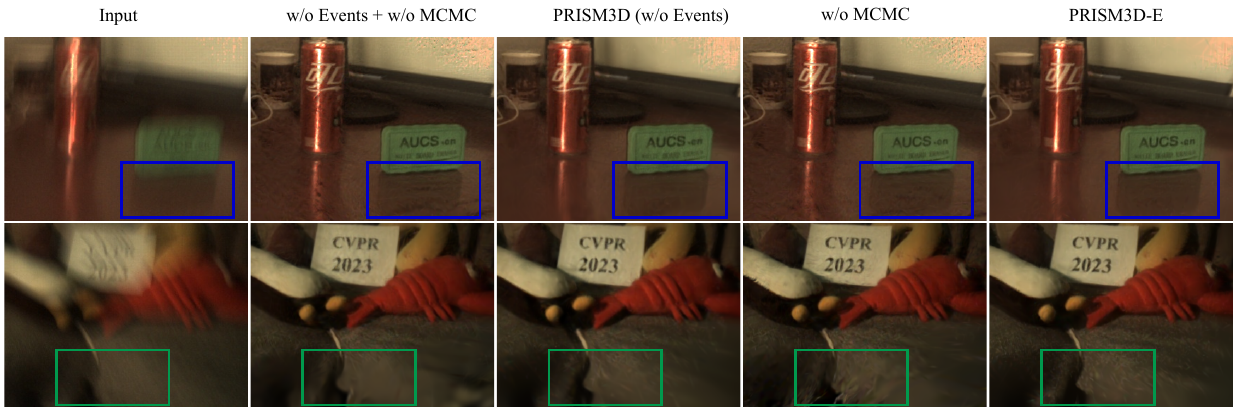}
    \caption{\textbf{Qualitative ablation study on the Real Dataset.} Consistent with our synthetic findings, \textbf{PRISM3D} produces robust reconstructions from severe real-world motion blur, relying on MCMC to effectively suppress optimization artifacts. Incorporating event priors in \textbf{PRISM3D-E} further refines the geometry and textures, yielding the most visually consistent and artifact-free outputs.}
    \label{fig:ablations_real}
\end{figure*}

To visually validate our quantitative findings, we present qualitative ablation results for both the synthetic and real-world datasets in Figure~\ref{fig:ablations_synthetic} and Figure~\ref{fig:ablations_real}, respectively. These visual comparisons demonstrate the specific impact of our core components on the final rendering quality. Specifically, we observe that the MCMC module is essential for suppressing floaters and background artifacts within our base RGB framework (PRISM3D). Furthermore, integrating complementary event data (PRISM3D-E) significantly enhances the recovery of high-frequency structural details, such as fine text, leading to the sharpest and most artifact-free reconstructions across both domains.

\section{Visualization of Reconstructed Scene Geometry}
\label{sec:extended_depth_visualization}

To further evaluate the quality of the reconstructed 3D scene geometry, we present additional visualizations of the rendered views together with their corresponding depth maps for five representative real-world scenes. These visualizations complement the quantitative evaluations by illustrating the structural consistency of the recovered scene representation. The estimated depth maps demonstrate that both \textbf{PRISM3D} and \textbf{PRISM3D-E} recover coherent scene geometry in addition to producing high-quality rendered views, allowing readers to assess reconstruction quality beyond appearance alone.

\begin{figure*}[!ht]
\centering

\includegraphics[width=\textwidth]{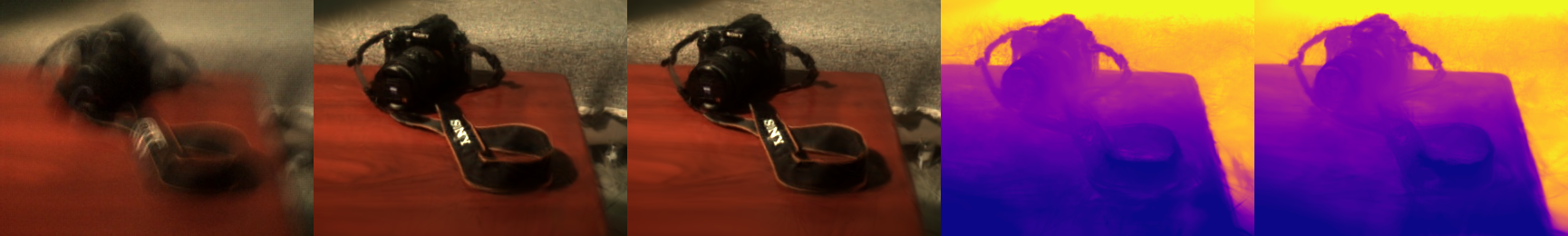}\\[0.5mm]
\includegraphics[width=\textwidth]{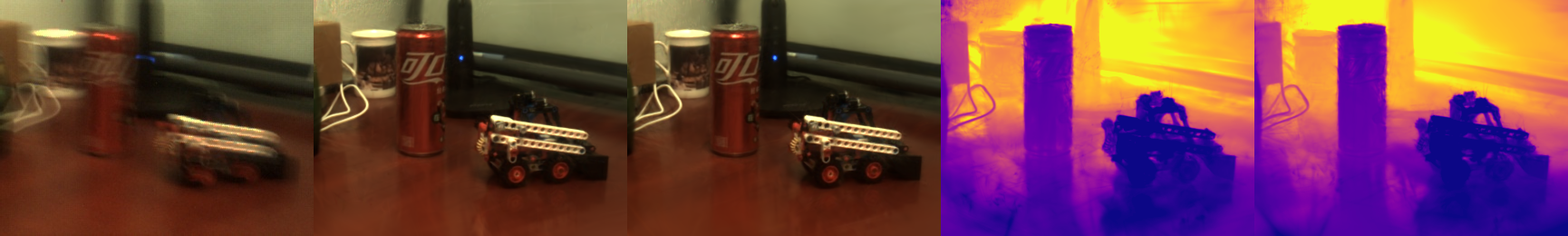}\\[0.5mm]
\includegraphics[width=\textwidth]{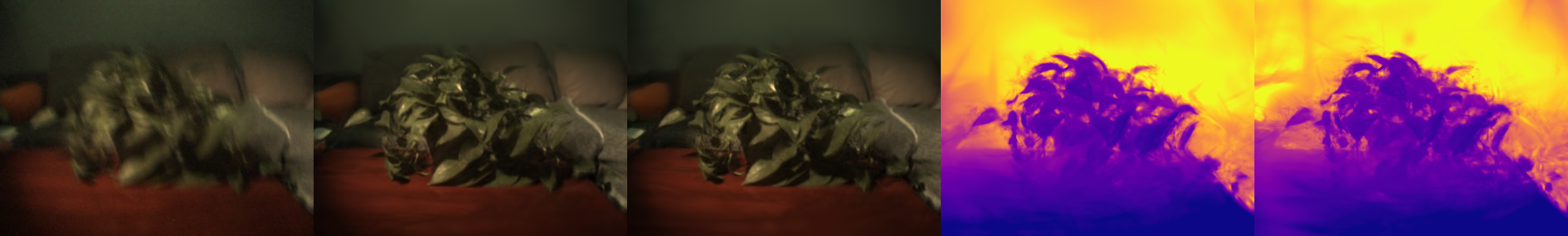}\\[0.5mm]
\includegraphics[width=\textwidth]{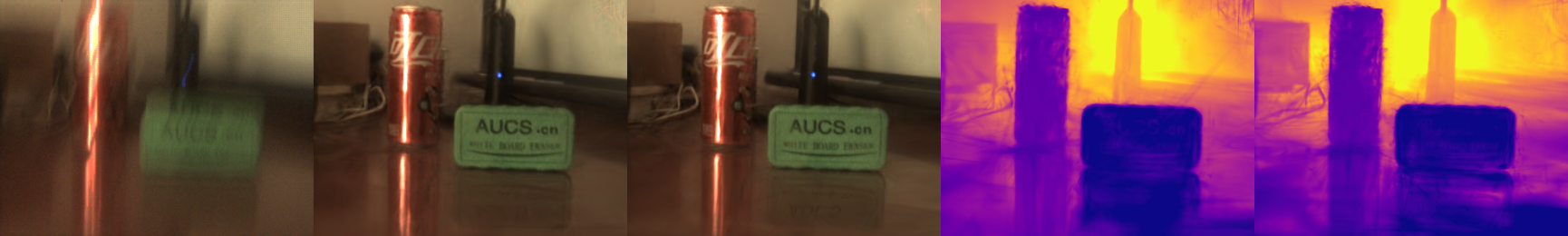}\\[0.5mm]
\includegraphics[width=\textwidth]{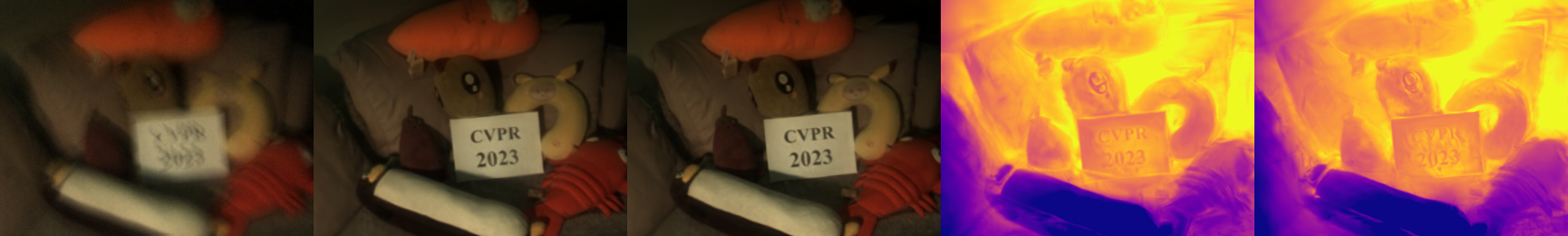}

\caption{
Visualization of the reconstructed scene geometry for five real-world scenes. Each row corresponds to a different scene. From left to right, the columns show: \textbf{(i)} input image, \textbf{(ii)} rendering produced by PRISM3D, \textbf{(iii)} rendering produced by PRISM3D-E, \textbf{(iv)} depth map estimated by PRISM3D, and \textbf{(v)} depth map estimated by PRISM3D-E. The rendered views together with the corresponding depth maps demonstrate that both methods recover coherent scene structure in addition to producing high-quality novel view synthesis.
}

\label{fig:depth_comparison}
\end{figure*}

\clearpage

\section{Extended Visualizations of PRISM3D}
\label{sec:extended_visualizations_appendix}

Beyond the comparative qualitative evaluations, we provide extended visual galleries to demonstrate the deblurring robustness of \textbf{PRISM3D} and \textbf{PRISM3D-E}. Because extreme motion blur varies significantly across different views within a scene, we showcase restoration results on multiple distinct frames (e.g., Frame A and Frame B). This confirms that our framework consistently recovers sharp textures and suppresses artifacts across diverse, complex blur patterns in both synthetic and real-world datasets.

\subsection{Synthetic Dataset Restorations}

\begin{figure*}[!hbp]
    \centering
    \includegraphics[width=0.72\linewidth]{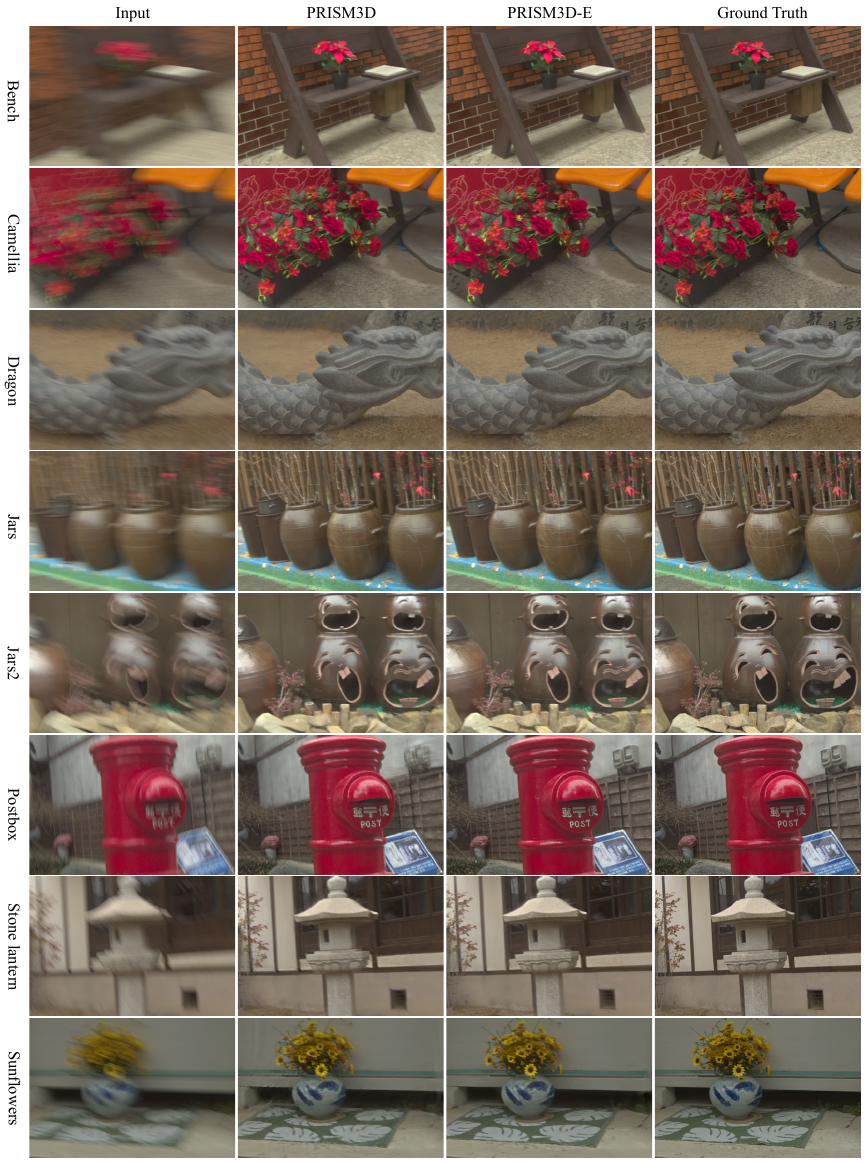}
    \caption{\textbf{Extended Visualizations on the Synthetic Dataset (Frame A):} Our methods consistently restore sharp, high-frequency details from severely blurred inputs. PRISM3D effectively resolves the dynamic blur, while PRISM3D-E further refines fine structures and text using complementary event priors.}
    \label{fig:Our_Synethetic_Results_1_Supplementary}
\end{figure*}
\clearpage

\begin{figure*}[!hbp]
    \centering
    \includegraphics[width=0.8\linewidth]{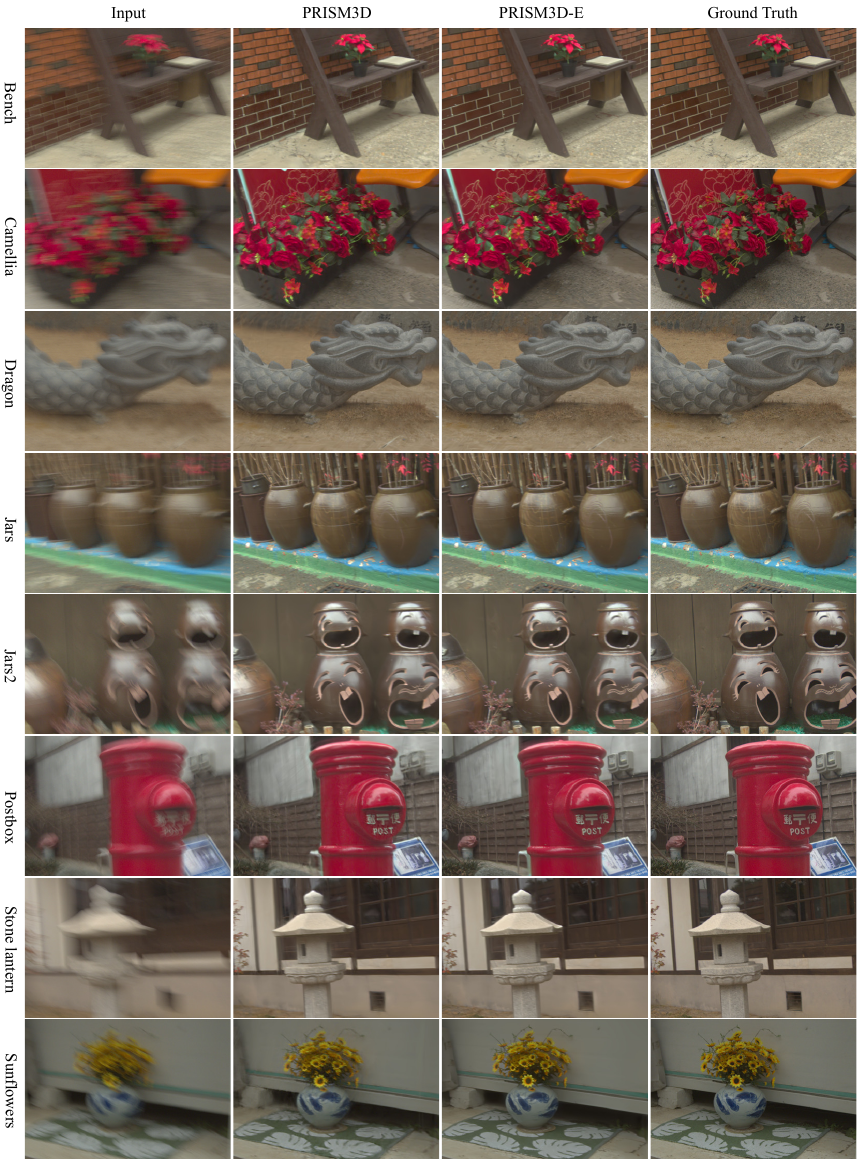}
    \caption{\textbf{Extended Visualizations on the Synthetic Dataset (Frame B):} Deblurring results on a different frame from the same synthetic sequences. This demonstrates that our framework's restoration quality remains stable and artifact-free across varying degrees of motion blur within the same trajectory.}
    \label{fig:Our_Synethetic_Results_2_Supplementary}
\end{figure*}
\clearpage

\subsection{Real Dataset Restorations}

\begin{figure*}[!hbp]
    \centering
    \includegraphics[width=0.9\linewidth]{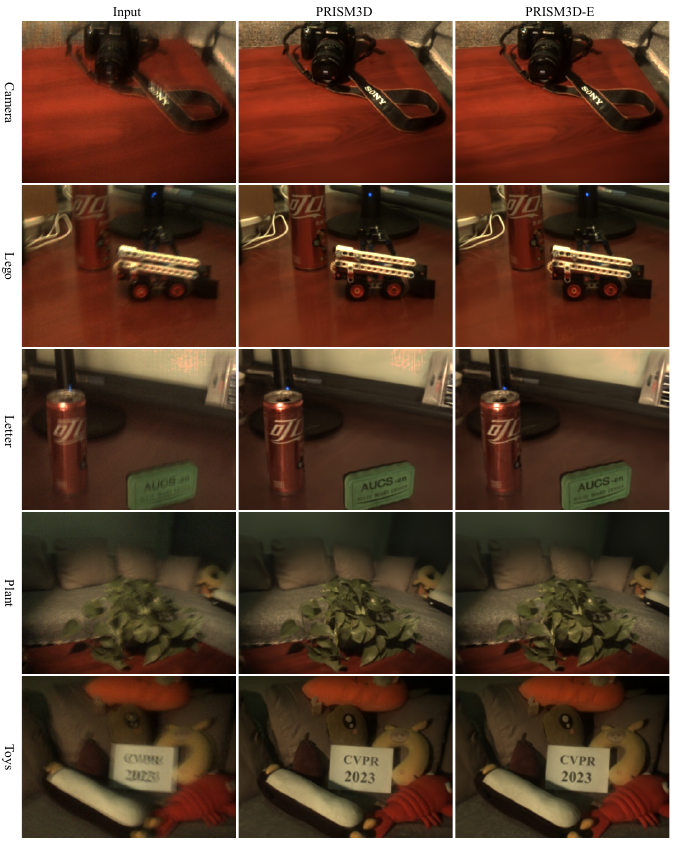}
    \caption{\textbf{Extended Visualizations on the Real Dataset (Frame A):} Real-world sequences exhibit highly complex, non-linear blur. PRISM3D successfully suppresses these severe degradations, and PRISM3D-E leverages event data to recover crisp textures that are otherwise permanently lost in the RGB exposure.}
    \label{fig:Our_Real_Results_1_Supplementary}
\end{figure*}
\clearpage

\begin{figure*}[!hbp]
    \centering
    \includegraphics[width=0.9\linewidth]{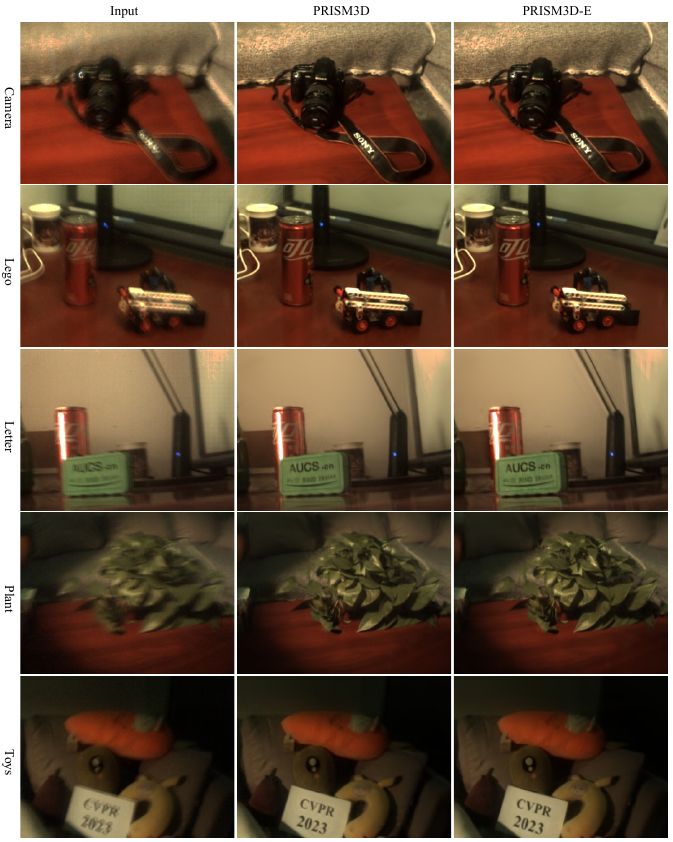}
    \caption{\textbf{Extended Visualizations on the Real Dataset (Frame B):} Restoration results on another distinct frame from the real-world captures, further confirming the temporal consistency and robustness of our physical blur modeling over the entire captured sequence.}
    \label{fig:Our_Real_Results_2_Supplementary}
\end{figure*}

\end{document}